%% file: main.tex
\pdfoutput=1

\RequirePackage[table,xcdraw]{xcolor}

\documentclass[11pt]{article}

\usepackage[preprint]{acl}

\usepackage{times}
\usepackage{latexsym}

\usepackage[T1]{fontenc}

\usepackage[utf8]{inputenc}
\usepackage{microtype}       
\usepackage{inconsolata}     

\usepackage{graphicx}        
\usepackage{wrapfig}         
\usepackage{float}           
\usepackage{adjustbox}       
\usepackage{booktabs}        
\usepackage{tabularx}        
\usepackage{makecell}        
\usepackage{multirow}        
\usepackage{arydshln}        
\usepackage{hhline}          

\definecolor{goldenyellow}{RGB}{240, 180, 45} 

\usepackage{enumitem}        
\usepackage{algpseudocode}   

\usepackage{amssymb}         
\usepackage{bbding}          
\usepackage{soul}            
\usepackage{tcolorbox}       

\usepackage{hyperref}        

\usepackage{pdfpages}        

\usepackage{fontawesome}     

\title{UrbanVideo-Bench: Benchmarking Vision-Language Models on Embodied Intelligence with Video Data in Urban Spaces}

\author{
    \textbf{Baining Zhao}$^{*}$,
    \textbf{Jianjie Fang}$^{*}$,
    \textbf{Zichao Dai}$^{*}$,
    \textbf{Ziyou Wang},
    \textbf{Jirong Zha},\\
    \textbf{Weichen Zhang}, 
    \textbf{Chen Gao}$^{\dagger}$,
    \textbf{Yue Wang},
    \textbf{Jinqiang Cui},
    \textbf{Xinlei Chen}$^{\dagger}$, 
    \textbf{Yong Li} \\
    Tsinghua University\\ \texttt{chen.xinlei@sz.tsinghua.edu.cn}, \texttt{chgao96@gmail.com}, \texttt{liyong07@tsinghua.edu.cn} \\
    \textcolor{magenta}{\faGlobe\ \href{https://embodiedcity.github.io/UrbanVideo-Bench/}{\textcolor{magenta}{\textit{Project Page}}}} \hspace{1cm}
    \textcolor{goldenyellow}{\faDatabase\ \href{https://huggingface.co/datasets/EmbodiedCity/UrbanVideo-Bench}{\textcolor{goldenyellow}{\textit{Dataset}}}} \hspace{1cm}
    \textcolor{black}{\faGithub\ \href{https://github.com/EmbodiedCity/UrbanVideo-Bench.code}{\textcolor{black}{\textit{Code}}}}
}

\begin{document}

\maketitle

\setlength{\abovecaptionskip}{2pt}   
\setlength{\belowcaptionskip}{-10pt}  

\begin{abstract}
    Large multimodal models exhibit remarkable intelligence, yet their embodied cognitive abilities during motion in open-ended urban 3D space remain to be explored. We introduce a benchmark to evaluate whether video-large language models (Video-LLMs) can naturally process continuous first-person visual observations like humans, enabling recall, perception, reasoning, and navigation. We have manually control drones to collect 3D embodied motion video data from real-world cities and simulated environments, resulting in 1.5k video clips. Then we design a pipeline to generate 5.2k multiple-choice questions. Evaluations of 17 widely-used Video-LLMs reveal current limitations in urban embodied cognition. Correlation analysis provides insight into the relationships between different tasks, showing that causal reasoning has a strong correlation with recall, perception, and navigation, while the abilities for counterfactual and associative reasoning exhibit lower correlation with other tasks. We also validate the potential for Sim-to-Real transfer in urban embodiment through fine-tuning. 
\end{abstract}

\input{1.Introduction}

\input{4.Related_Work}

\input{2.Benchmark}

\input{3.Experiments}

\input{5.Conclusion}

\bibliography{ref}

\appendix
\input{Appendix}

\end{document}

%% file: 1.Introduction.tex
\section{Introduction}

\begin{figure*}[h]
	\centering
	\includegraphics[width = \linewidth]{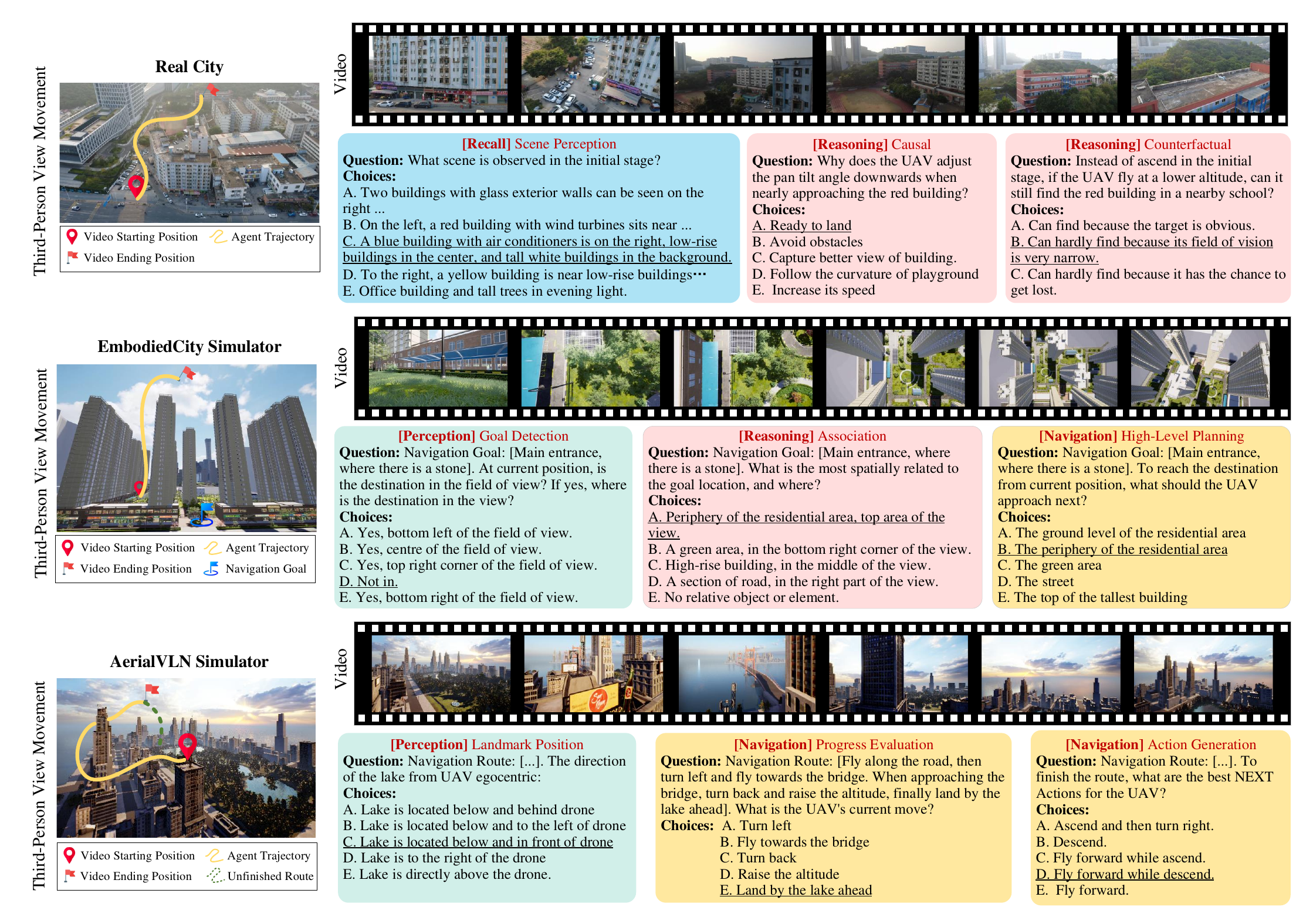}
    \vspace{-20pt}
	\caption{Example of video-language multiple choice question-answering. This figure presents three video examples corresponding to three data sources: real cities in Guangdong Province, China; the simulator EmbodiedCity constructed based on the real city of Beijing, China; and the simulator AerialVLN built on virtual cities. To ensure logical consistency in the movement trajectories within the videos, all video clips consist of continuous perceptual observations generated from ongoing or completed vision-language navigation tasks. For each video clip, we designed different task types to evaluate the embodied cognitive intelligence of Video-LLMs.}
    \label{Fig:fig2.1}
    \vspace{-5pt}
\end{figure*}

\begin{table*}[h]
	\centering
	\caption{Task set overview. Embodied cognition in motion is divided into four abilities, each of which is manifested in several tasks. Each task is provided with corresponding handcrafted question prototypes.}
	\renewcommand{\arraystretch}{1.1}
    \begin{small} 
		\begin{tabularx}{\textwidth}{lX}
			\hline
			\rowcolor{cyan!20} \multicolumn{2}{c}{\textbf{Recall}}\\ \hline
			\textbf{Trajectory Captioning} & Summarize the agent's movement path using visual cues / landmarks. \\ \hline
			\textbf{Sequence Recall} & What is the agent's next step after [changing direction to the left over the intersection]? \\ \hline
			\textbf{Object Recall} & what is [located next to] the [Central ALL-STAR cafe]?  \\ \hline
			\textbf{Scene Recall} & Describe scene the agent observes [during the descent to a lower height near the destination].  \\ 
            \hline
            \textbf{Start/End Position} & Where are the starting point and final destination of the agent's movement? \\ \hline
			\rowcolor{green!20} \multicolumn{2}{c}{\textbf{Perception}}   \\ \hline
			\textbf{Proximity} & How does the distance between the agent and the [rooftop with solar panels in the residential area] change after the [agent descends to street level]?  \\ \hline
			\textbf{Duration} & Which takes longer, [the agent's movement through the skyscraper alley] or [its descent to the balcony on the 9th floor]?  \\ \hline
			\textbf{Landmark Position} & Given [navigation goal/route] initially, what is agent's current position relative to [landmark]?  \\ \hline
			\textbf{Goal Detection} & Given [navigation goal] at starting location, is the target currently visible in the field of view, and if so, what is its position within the view?\\ \hline
			\textbf{Cognitive Map} & Summarize historical movement observations into a cognitive map.   \\ \hline
			\rowcolor{red!20} \multicolumn{2}{c}{\textbf{Reasoning}}\\ \hline
			\textbf{Causal} & Why did the agent [perform a descent after ascending alongside the cylindrical building]?  \\ \hline
			\textbf{Counterfactual} & Instead of [flying over the elevated highway intersection], if the agent chooses to [fly around the cylindrical building], can it complete the task, and how is the alternative route?  \\ \hline
			\textbf{Association} & Given [navigation goal] at starting location, are there any relevant urban elements or objects in sight when the navigation goal is not visible?  \\ \hline
			\rowcolor{orange!20} \multicolumn{2}{c}{\textbf{Navigation}}\\ \hline
			\textbf{Progress Evaluation} &  (Route-oriented vision-language navigation) Given [navigation route] at starting location, analyze which step the navigation is currently perform. \\ \hline
			\textbf{High-level Planning}& (Goal-oriented vision-language navigation) Given [navigation goal] at starting location, make next plan from the current location.  \\ \hline
			\textbf{Action Generation}& Given [navigation goal/route] initially, generate the next control action from the current location.  \\ \hline
		\end{tabularx}
        \end{small} 
	\label{table_task_set}
    \vspace{-10pt}
\end{table*}

Humans can process continuous first-person visual observations, enabling them to discern direction, judge distance, and navigate in the three-dimensional space of the real world \cite{richardson2010challenging,burigat2017mobile,grauman2022ego4d,song2023llm}.  
This refers to embodied cognition in motion, which highlights that cognitive processes are deeply rooted in the body's interactions with the world~\cite{shapiro2019embodied,newcombe2023building}.
Naturally, endowing agents with this embodied cognition capability has been a long-term goal in the field of embodied intelligence \cite{fan2022minedojo,singh2023scene}.

In recent years, large multimodal models (LMMs) \cite{li2023large,wang-etal-2024-mm,xu2025defining} have emerged as a promising approach to achieve this goal. Typically, video-large language models (Video-LLMs) are evaluated on capabilities such as video summarization \cite{samel2024exploring,hua2024v2xum}, event question answering \cite{wang2024weakly}, and goal localization \cite{yu2023mm}. However, the benchmarks used to assess these capabilities are often limited to disembodied third-person video clips, where the agent itself is static~\cite{wu2024longvideobench,song2024moviechat,fang2024mmbench}. Besides, existing embodied video understanding research \cite{suglia2024alanavlm,cheng2024videgothink} mainly focuses on robotic arm manipulation~\cite{nair2022r3m} or indoor/ground-level movement~\cite{marcu2024lingoqa,yang2024thinking}. However, the embodied cognitive abilities required for three-dimensional motion in urban open-ended spaces have not been well-defined or assessed.
The first-person perspective visual continuous observations generated in this scenario possess the following characteristics:

\begin{itemize}[leftmargin=*]
	\item Complex Scene and Rich Semantic Information: Urban areas are vast, containing diverse elements like skyscrapers, bridges, and tunnels that provide rich semantic information and pose comprehension and navigation challenges, while dynamic elements like pedestrians and vehicles require real-time adaptation~\cite{yao2024aeroverse, xu2023urban}.
	\item Unique Aerial Motion: Aerial navigation involves vertical mobility and a first-person perspective, adding complexity by requiring enhanced embodied cognition for processing diverse motion and observation angles, necessitating advanced spatial awareness and decision-making~\cite{gao2024embodiedcity, lee2024citynav}.
	
\end{itemize} 
From these characteristics, we infer that embodied cognition in urban open spaces poses new challenges, and assessing LMMs' embodied cognitive abilities offers insights for future urban applications~\cite{wang20248th}.

We can use drones to capture motion video in urban spaces as they navigate buildings and dynamic elements. Establishing a benchmark presents challenges: 1) Creating a task set to evaluate embodied capabilities in urban spaces. 2) Obtaining video data: Unlike most high-altitude aerial views~\cite{li2016multi,zhu2021detection,dronecrowdcvpr2021}, our goal is to record drones maneuvering among urban structures with flexible movement and camera angles. Drones face issues like signal loss, limited range, and crashes due to obstructions and interference, making data collection difficult and costly. 3) Designing logical and purposive motion routes to ensure coherent visual observations.

Accordingly, we introduce a benchmark, \textbf{UrbanVideo-Bench}, designed for embodied motion cognition from embodied videos in urban airspace. 
Firstly, we propose a novel task set comprising 16 tasks characterized by urban spatiotemporal features, as shown in Figure \ref{Fig:fig2.1} and Table \ref{table_task_set}.
Secondly, we manually operate drones to collect embodied video data from 1) the real cities in Guangdong Province, China, 2) a simulator EmbodiedCity~\cite{gao2024embodiedcity} built on the real city Beijing, China, and 3) a simulator AerialVLN~\cite{liu2023aerialvln} built on virtual cities. Using both real devices and simulators helps to rapidly increase the number of videos. The movements in the videos are intentional, directed towards navigating to a specific position within urban space or following a particular route~\cite{wu2024vision}. 
Then, we developed a question-answer generation pipeline with trained human annotators (over 800 hours of effort) and the expertise of LMMs, generating high-quality multiple-choice questions (MCQs). 
Finally, we quantitatively and qualitatively evaluate widely-used Video-LLMs in zero-shot settings, including both proprietary and open-source models. We additionally attempt supervised fine-tuning (SFT) on two Video-LLMs to validate the effectiveness of our dataset.

Overall, the innovation of this research is the establishment of \textbf{the first benchmark for embodied cognition specifically tailored for motion in urban open-ended spaces}:

\begin{itemize}[leftmargin=*]
	\item We propose a novel task set comprising 4 categories and 16 tasks to evaluate how Video-LLMs recall, perception, reasoning, and navigation from embodied videos.
	\item We consequently develop 5.2k multiple-choice questions and 1.5k video clips, derived from real world and simulated environments. The dataset generation pipeline can be extended to other embodied movement videos.
	\item 17 popular LMMs are evaluated and their shortcomings are analyzed. We also explored the correlation between embodied cognitive abilities and the potential of Sim-to-Real.
	
\end{itemize}

%% file: 4.Related_Work.tex
\section{Related Work}

\begin{table*}[ht]
\centering
\caption{The proposed and popular benchmarks for video-large language models. Our benchmark's video sources and scenarios are different from others, focusing on evaluating the embodied cognitive abilities of Video-LLMs related to urban 3D aerial motion.}
\label{tab:datasets}
\scalebox{0.62}
{
\begin{tabular}{l|l|l|c|l|l|c|c}
\hline
\textbf{Benchmark} & \textbf{Video Source} & \textbf{Video Theme} & \textbf{Embodied} & \textbf{Environment} & \textbf{Motion} & \textbf{Video Num.} & \textbf{QA Num.} \\ 
\hline
LongVideoBench~\cite{wu2024longvideobench} & Web-Collected & Life, Movie, News & $\times$ & / & / & 3.8k & 6.7k \\
MovieChat-1K~\cite{song2024moviechat} & Web-Collected & Movie, TV series & $\times$ & / & / & 1.0k & 14.0k \\ 
MMBench-Video~\cite{fang2024mmbench} & YouTube & Life & $\times$ & / & / & 600 & 2.0k \\ 
HourVideo~\cite{chandrasegaran2024hourvideo} & Public Dataset & Human Activity & $\checkmark$ & Indoor/Outdoor & Ground-Level & 500 & 13.0k \\  
EgoSchema~\cite{mangalam2023egoschema} & Public Dataset & Human Activity & $\checkmark$ & Indoor/Outdoor & Ground-Level & 5.0k & 5.0k \\
Lingoqa~\cite{marcu2024lingoqa} & Self-Recorded & Driving & $\checkmark$ & Outdoor & Ground-Level & 28.0k & 419.0k \\
VSI-Bench~\cite{yang2024thinking} & 3 Public Datasets & Indoor Motion & $\checkmark$ & Indoor & Ground-Level & 288 & 5.0k \\
\hline
Ours & Self-Recorded & Aerial Agent Motion & $\checkmark$ & City & 3D Aerial Space & 1.5k & 5.2k \\ 
\hline
\end{tabular}
}
\vspace{-10pt}
\end{table*}

\textbf{Embodied Capabilities of LMMs.} Embodied intelligence refers to the concept that cognitive processes are deeply rooted in the body's interactions with the world~\cite{gupta2021embodied,shi2024opex,ding2024understanding,zeng2024perceive}. Large Multimodal Models (LMMs) have demonstrated unprecedented visual understanding capabilities and are considered the "brains" for developing embodied agents~\cite{tang2023video,liang2024survey,huang2023embodied}. Unlike past work~\cite{du2024embspatial,ramakrishnan2024does} that primarily focused on 2D images, static point cloud or language-based spatial understanding, human comprehension of the world is grounded in continuous visual perception~\cite{zhang2024navid,liao2024videoinsta,majumdar2023we}, akin to embodied cognition through video streams. Thus, we need relative video benchmarks from diverse sources to comprehensively evaluate the potential of LMMs in various embodied scenarios.

\textbf{Video Benchmarks for LMMs.} Traditional video benchmarks cover various tasks like abstract understanding and spatiotemporal analysis~\cite{xu2017video,wu2024star,fu2024video}. They mainly concentrate on understanding the video content~\cite{wu2024longvideobench,song2024moviechat,fang2024mmbench}, lacking exploration of Video-LLMs' embodied cognitive abilities from an embodied, egocentric perspective. While some research has focused on embodied capabilities in indoor or ground-level scenes~\cite{chandrasegaran2024hourvideo,mangalam2023egoschema,sima2024drivelm,marcu2024lingoqa,yang2024thinking}, there has been insufficient exploration of embodied abilities in urban open 3D spaces. Part of above-mentioned video benchmarks are shown in Table \ref{tab:datasets}. Comparatively, we independently record embodied motion video data along with corresponding MCQs to evaluate models' cognitive abilities in complex, dynamic urban 3D spaces.

%% file: 2.Benchmark.tex
\section{Benchmark Design and Construction}

We firstly define 16 embodied tasks that assess the embodied cognitive capabilities of Video-LLMs from different four aspects. Then, we describe the dataset generation process. Finally, we provide the statistical characteristics of the dataset.

\subsection{Task Set}

Considering the characteristics of urban open-ended scenarios, embodied cognition in motion can be divided into four abilities: recall, perception, reasoning, and navigation~\cite{chen2011collecting,tang2021human,chandrasegaran2024hourvideo}. Each ability is evaluated through several specific tasks, which are outlined in Table \ref{table_task_set}.

\textbf{Recall} tasks evaluates Video-LLM ability to cognitively remember key aspects of urban environments. By integrating the city's semantic elements seen in the video, the task includes Trajectory Captioning, where agents summarize their paths using landmarks. Sequence Recall asks agents to sort out the sequence of actions, while Object Recall focuses on identifying objects near landmarks. Scene Recall involves describing details observed during specific actions, ensuring comprehensive memory and understanding in dynamic urban contexts. The "Start/End Position" indicates whether agents are aware of "where I come from" and "where I am."

\textbf{Perception} include static relative spatial relationships (Landmark Position, Goal Detection), dynamic position changes (Proximity), temporal understanding (Duration), and scene-level comprehension (Cognitive Map). The design of these tasks encompasses a range from static to dynamic, spatial to temporal, and micro to macro levels.

\textbf{Reasoning} focuses on analyzing and making sense of its actions and surroundings within the urban environment. This includes understanding the causal relationships behind movements, such as why the agent descends after ascending alongside a building. It also involves counterfactual reasoning, where agents consider alternative routes, like flying around a building instead of over a highway, and assess the viability of these options. Additionally, Video-LLMs are required to recognize associations between visible urban elements and their navigation goals, even when the goals themselves are not in direct view, showcasing urban commonsense and the ability to think critically and adaptively.

\textbf{Navigation} tasks aims to evaluate whether Video-LLMs can plan routes and directly output actions in urban spaces~\cite{wu2024vision}. We assessed two types of vision-language navigation (VLN): route-oriented VLN~\cite{zhou2024navgpt}, where the agent is provided with a navigation route (e.g., fly forward to the white building, then turn right, and continue to the lakeside to stop) and can assess the progress of the current route (Progress Evaluation) while ultimately outputting control actions (Action Generation); and goal-oriented VLN~\cite{chaplot2020object}, where only a navigation goal (e.g., the lakeside) is given, allowing the agent to autonomously perform high-level navigation planning (High-level Planning) and eventually map it into control actions (Action Generation). Three-dimensional aerial navigation in urban environments is one of the most challenging tasks in embodied intelligence.

The proposed tasks reflect real-world challenges that embodied video systems in urban open-ended spaces may encounter, enhancing the practical relevance of the evaluation.

\subsection{Dataset Generation Pipeline}

\begin{figure*}[t]
	\centering
	\includegraphics[width = 0.99\linewidth]{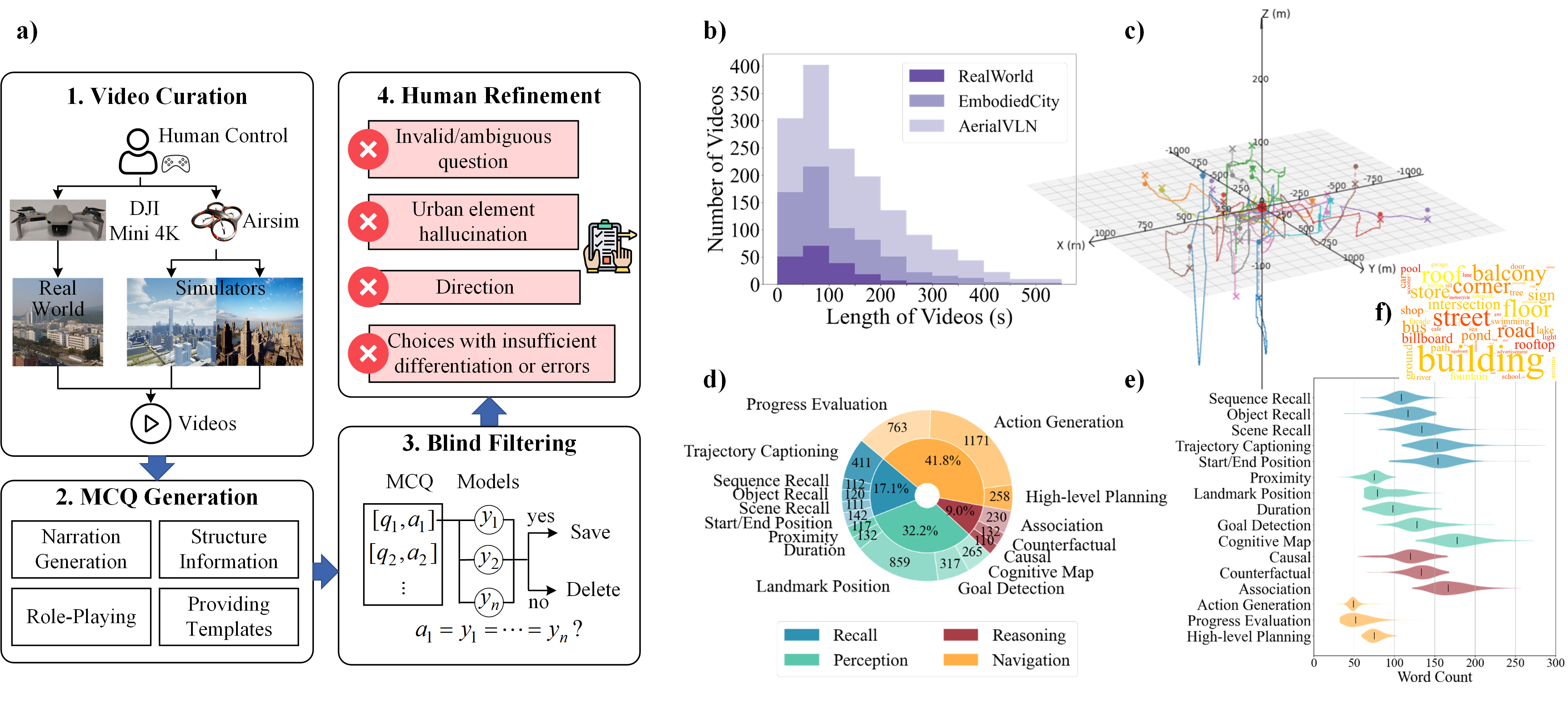}
	\caption{\textbf{Dataset Generation Pipeline and Statistics.} a) The pipeline includes four steps: video curation, MCQ generation, blind filtering, and human refinement. b) Histogram of frame count of videos. c) Histogram of path lengths. d) Histogram of word count of questions. e) Violin plot of word count of different categories of questions. f) Word cloud generated from questions and choices.}
	\label{Fig:fig2.3}
\end{figure*}

As presented in Figure \ref{Fig:fig2.3}a, we will outline the pipeline developed for creating the dataset, which primarily consists of four steps: video curation, multiple-choice
question-answering (MCQ) generation, blind filtering, and human refinement.

\subsubsection{Video Curation}

The primary consideration for UrbanVideo-Bench is obtaining massive high-quality embodied mobility video data. We collecte drone flight video data in two cities, Shenzhen and Zhaoqing, in Guangdong Province, China. The data collection was conducted using two DJI Mini 4K drones. To further expand the dataset, we chose to acquire data from simulators with two city benchmarks: EmbodiedCity~\cite{gao2024embodiedcity} and AerialVLN \cite{liu2023aerialvln}. They both have the following advantages:
a) Realistic environment modeling; b) Support for aerial agents; c) Existing aerial route reference. 

We divided our team into two groups. The first group designs the navigation goal/route and provides text instructions for the endpoint, ensuring that the final destination can be reached based on the instructions and common urban knowledge. The second group, consisting of experienced pilots (with over 1000 hours of real drone flight time), performs flights to perform the VLN tasks, collecting first-person perspective data during the flight. Through this approach, the movement of the agent is purposeful, and the collected video data is more conducive to evaluating the capabilities of Video-LLMs. See Appendix \ref{appendix:video_curation} for details.

\subsubsection{MCQ Generation}

The objective of this phase is to leverage the foundational video understanding and text generation capabilities of a powerful LMM to quickly produce high-quality MCQs for each task. We have designed a Chain-of-thought (CoT) prompting based on characteristics of embodied movement videos, consisting of the following four steps:
a) Narration Generation;
b) Structured Extraction of Key Information;
c) Role-playing;
d) Providing Question/Option Templates and Examples.
The detail process and prompts are in Appendix \ref{appendix:MCQ_generation}. In order to improve the quality of problem generation, we use Gemini-1.5 Flash \cite{team2024gemini} with video understanding capability to assist in generation at this stage.

\subsubsection{Blind Filtering}
The purpose of blind filtering is to eliminate MCQs that can be answered correctly based solely on urban common sense, without any video input. This process aims to evaluate the embodied cognition capabilities of Video-LLMs based on mobile video, rather than their world knowledge or common sense. Blind filtering enhances the quality of the dataset. The specific method involves using $n$ Video-LLMs to guess the answer to any given MCQ case without video input. If all $n$ Video-LLMs guess correctly, the MCQ is removed; otherwise, it is retained.

\subsubsection{Human Refinement}

The generated MCQs may contain invalid questions, ambiguous options, incorrect answers, and various other issues that require further human refinement. These issues stem from two main sources: a) Even the most advanced Video-LLMs lacks the ability to fully understand embodied movement videos. b) The urban aerial agent scenarios are complex, making video comprehension challenging. We approach the human refinement process from four aspects: a) Invalid/ambiguous questions: For example, when the task specifies for the agent to "navigate to the main entrance," the presence of multiple nearby buildings leads to ambiguity, as the agent is unsure which building's main entrance is intended. The navigation target should be clarified to "navigate to the main entrance of the yellow building on the right." b) Urban element hallucination: This refers to the presence of city elements or objects in the correct option that were never actually present in the video. c) Direction: Directional descriptions are often incorrect or imprecise. d) Choices with insufficient differentiation or errors: One scenario is more than one option is correct. Another is only one is correct, but the ground truth was provided incorrectly. The entire refinement process required over 800 person-hours. See Appendix \ref{appendix:human_refinement} for more examples.

\subsection{Dataset Statistics}

The proposed dataset includes 1,547 video clips with resolutions of 1280x720 pixels for real drone frames, 960x720 pixels for EmbodiedCity frames, and 520x520 pixels for AerialVLN frames.
As shown in Figure \ref{Fig:fig2.3}b\&c, the durations of these video clips range from 10 seconds to 10 minutes, and the trajectories of the UAV in the videos encompasses different directions and movement patterns in 3-dimensional spaces, thus suggesting our dataset covers diverse scenarios from brief fine movements, medium-range command execution to long-distance navigation. The dataset also possesses over 5.2K multi-choice questions, including low-level tasks such as Recall and Perception, and also high-level tasks such as Reasoning and Navigation. Seen from Figure \ref{Fig:fig2.3}d, both low-level and high-level tasks take up to approximately 50\% of the total questions, enabling a comprehensive evaluation of the various capabilities required for Video-LLMs to perform embodied cognitive tasks during motion in urban open-ended spaces. For each question, the word count ranges from 50 to 250, varying based on the question category and the complexity of the environment and tasks included in the specific video, as is shown in Figure \ref{Fig:fig2.3}e. We finally generate a word cloud in Figure \ref{Fig:fig2.3}f, demonstrating the richness of urban elements included in our MCQs.

%% file: 3.Experiments.tex
\section{Experiments}

\setlength{\tabcolsep}{4pt} 
\begin{table*}[t]
	\scriptsize{
	\centering
        \caption{Accuracy of different Video-LLMs on embodied tasks. The former section shows existing popular Video-LLMs' results. The latter section demonstrates fine-tuning results for two models, highlighting sim-to-real potential.}
    \label{tab:example_table_large_font}
	\renewcommand{\arraystretch}{1.1}
        \scalebox{0.96}
        {
	\begin{tabular}{r|cc|ccccccccllllllll}
	\hline
	& \multicolumn{1}{l}{}                         
	& \multicolumn{1}{l|}{}                         
	& \multicolumn{5}{c}{\cellcolor{cyan!20}Recall}                   
	& \multicolumn{5}{c}{\cellcolor{green!20}Perception}               
	& \multicolumn{3}{c}{\cellcolor{red!20}Reasoning}                
	& \multicolumn{3}{c}{\cellcolor{orange!20}Navigation}               \\Method                                                               & \multicolumn{1}{l}{Rank}   & \multicolumn{1}{l|}{Avg.}              & \rotatebox{90}{\textit{Trajectory Captioning}}                       
	& \rotatebox{90}{\textit{Sequence Recall}  }                    
	& \rotatebox{90}{\textit{Object Recall}   }                    
	& \rotatebox{90}{\textit{Scene Recall} }                       
	& \rotatebox{90}{\textit{Start/End Position} }                 
	& \rotatebox{90}{\textit{Proximity}     }                      
	& \rotatebox{90}{\textit{Duration}}                            
	& \rotatebox{90}{\textit{Landmark Position}}                   
	& \multicolumn{1}{c}{\rotatebox{90}{\textit{Goal Detection}}} 
	& \multicolumn{1}{c}{\rotatebox{90}{\textit{Cognitive Map}}} 
	& \multicolumn{1}{c}{\rotatebox{90}{\textit{Causal}}} 
	& \rotatebox{90}{\textit{Counterfactual}} 
	& \rotatebox{90}{\textit{Association} } 
	& \rotatebox{90}{\textit{Progress Evaluation} }
	& \rotatebox{90}{\textit{High-level Planning} }
	& \rotatebox{90}{\textit{Action Generation}} \\ \hline
	\rowcolor[HTML]{ECF4FF} 
		\multicolumn{1}{l|}{\cellcolor[HTML]{ECF4FF}\textit{Baseline}}     & \multicolumn{1}{l}{\cellcolor[HTML]{ECF4FF}} 
		& \multicolumn{1}{l|}{\cellcolor[HTML]{ECF4FF}} 
		& \multicolumn{1}{l}{\cellcolor[HTML]{ECF4FF}} 
		& \multicolumn{1}{l}{\cellcolor[HTML]{ECF4FF}} 
		& \multicolumn{1}{l}{\cellcolor[HTML]{ECF4FF}} 
		& \multicolumn{1}{l}{\cellcolor[HTML]{ECF4FF}} 
		& \multicolumn{1}{l}{\cellcolor[HTML]{ECF4FF}} 
		& \multicolumn{1}{l}{\cellcolor[HTML]{ECF4FF}} 
		& \multicolumn{1}{l}{\cellcolor[HTML]{ECF4FF}} 
		& \multicolumn{1}{l}{\cellcolor[HTML]{ECF4FF}} 
		& \multicolumn{1}{l}{\cellcolor[HTML]{ECF4FF}}
		& & & & & & &                            \\
	Random              & -     &19.7                               & 18.5  & 17.0  & 20.8 & 13.5  & 21.8                             & 37.8  & 35.6  & 19.7 & 18.0  & 21.9                                            
	& \multicolumn{1}{c}{18.2}                       
	&\multicolumn{1}{c}{25.0}                      
	& \multicolumn{1}{c}{18.3}               
	& \multicolumn{1}{c}{21.8}   
	& \multicolumn{1}{c}{15.9} 
	& \multicolumn{1}{c}{16.4}          \\ \hline
	\rowcolor[HTML]{ECF4FF} 
	\multicolumn{1}{l|}{\cellcolor[HTML]{ECF4FF}\textit{Proprietary Models (API)}} 
	&  &  &  &  &  &  &  &  &  &  &  &  &  &  &  &  &  &  \\
		Gemini-1.5-Flash[1 fps]                                                     & 4 & 40.5 & 39.7 & 51.8 & 61.7 & 79.3 & 61.3 & 47.1 & 59.8 
		  & 37.8 & 28.7 & 47.9 & 60.0 & 42.4 & 20.0 & 43.3 & 32.6 & 34.4\\
		Gemini-1.5-Pro[1 fps]                                                      &\cellcolor[HTML]{FFCCC9}{3} & 42.5 & 58.6 & 61.6 & 65.0 & 72.1 & 66.2 & 66.4 & 63.6 
		  & 37.4 & 33.8 & 46.0 & 63.6 & 46.2 & 23.0 & 38.8 & 43.8 & 31.9\\
		Gemini-2.0-Flash[1 fps]                                                     & 5 & 38.3 & 47.9 & 58.9 & 63.3 & 75.7 & 57.0 & 66.4 & 47.7  
		  & 27.9 & 27.8 & 45.3 & 62.7 & 24.2 & 17.8 & 39.2 & 48.4 & 30.5\\
		GPT-4o-mini[32f]                                                          & 6 & 36.5 & 33.0 & 53.6 & 48.3 & 59.5 & 56.3 & 69.7 & 51.5 
		  & 33.3 & 31.3 & 42.4 & 65.5 & 47.7 & 22.9 & 30.8 & 57.5 & 25.4\\
		GPT-4o[32f]                                                               &\cellcolor[HTML]{FD6864}{2}  & 43.6 & 47.6 & 58.9 & 65.0 & 67.6 & 61.3 & 63.0 & 47.7 
		  & 36.8 & 42.4 & 52.8 & 66.4 & 44.7 & 45.8 & 34.2 & 67.8 & 33.8\\
		Qwen-VL-Max-latest[32f]                                                   &\cellcolor[HTML]{FE0000}{1} & 45.5 & 44.9 & 70.5 & 64.2 & 75.7 & 73.9 & 78.2 & 43.9 
		  & 44.8 & 44.7 & 61.1 & 77.3 & 49.2 & 23.9 & 38.8 & 70.0 & 29.6\\ \hline
	\rowcolor[HTML]{ECF4FF} 
	\multicolumn{1}{l|}{\cellcolor[HTML]{ECF4FF}\textit{Open-source Models}}       
	& & & & & & & & & & & & & & & & & & \\
		LLaVA-NeXT-Video-7B-hf[32f]                                               &\cellcolor[HTML]{FFCCC9}{3} & 38.6 & 55.7 & 39.3 & 43.3 & 61.3 & 40.8 & 58.8 & 52.3  
		  & 49.5 & 16.7 & 26.8 & 44.5 & 20.5 & 58.7 & 36.6 & 52.3 &19.2\\
		Phi-3.5-vision-instruct[32f]                                             &\cellcolor[HTML]{FD6864}{2} & 38.7 & 67.0 & 57.1 & 57.5 & 64.9 & 45.1 & 48.7 & 45.5 
		  & 49.2 & 17.0 & 52.1 & 51.8 & 34.8 & 13.9 & 33.2 & 59.7 &15.6\\
		Kangaroo[64f]                                                             &\cellcolor[HTML]{FE0000}{1} & 39.2 & 27.0 & 66.1 & 60.8 & 69.4 & 53.5 & 75.6 & 57.6 
		  & 35.5 & 37.2 & 60.0 & 64.5 & 42.4 & 19.1 & 32.5 & 41.9 & 32.4\\
		Qwen2-VL-2B-Instruct[0.5 fps]                                                 & 5 & 31.9 & 29.9 & 54.5 & 30.8 & 57.7 & 24.6 & 69.7 & 47.7 
		  & 22.0 & 22.1 & 64.2 & 46.4 & 35.6 & 13.5 & 28.8 & 44.2 & 27.3\\
		Qwen2-VL-7B-Instruct[0.25 fps]                                                 & 4 & 36.2 & 36.5 & 50.9 & 47.5 & 65.8 & 47.2 & 52.1 
		  & 48.5 & 25.1 & 28.4  & 55.8 & 55.5 & 29.5 & 11.7 & 33.9 & 59.3 & 32.7 \\
		InternVL2-2B[32f]                                                          & 11 & 27.6 & 19.2 & 29.5 & 37.5 & 55.9 & 22.5 & 57.1 & 37.9 
		   & 19.3 & 24.6 & 39.2 & 33.6 & 45.5 &33.5 & 29.2 & 37.6 & 20.9\\
		InternVL2-4B[32f]                                                          & 10 & 28.1 & 19.2 & 37.5 & 33.3 & 62.2 & 24.6 & 66.4 & 42.4 
		   & 23.2 & 26.5 & 32.8 & 36.4 & 35.6 & 24.8 & 29.5 & 32.2 & 22.1\\
		InternVL2-8B[32f]                                                          & 9 & 28.1 & 23.4 & 23.2 & 35.0 & 52.3 & 22.5 & 58.0 & 44.7 
		   & 23.1 & 27.4 & 28.3 & 33.6 & 45.5& 27.0 & 31.5 & 35.7 & 21.4 \\
		InternVL2-26B[32f]                                                         & 8 & 28.3 & 24.3 & 36.6 & 35.0 & 61.3 & 26.8 & 51.2 & 40.2 
		   & 19.9 & 28.1 & 32.4 & 32.7 & 44.7 & 26.5 & 28.9 & 37.6 &22.8\\
		InternVL2-40B[32f]                                                         & 7 & 28.4 & 22.2 & 19.6 & 30.8 & 54.1 & 21.1 & 61.3 & 50.0 
		   & 23.2 & 26.5 & 34.7 & 27.3 & 41.7 & 25.7 & 32.4 & 34.9 & 22.3 \\
		InternVL2-Llama3-76B[32f]                                                  & 6 & 28.9 & 19.5 & 38.4 & 37.5 & 54.1 & 18.3 & 65.5 & 48.5 
		   & 22.9 & 28.1 & 33.6 & 30.9 & 43.2 & 27.4 & 31.3 & 34.5 & 23.2\\
		   \hhline{===================}
	\rowcolor[HTML]{ECF4FF} 
	\multicolumn{1}{l|}{\cellcolor[HTML]{ECF4FF}\textit{Fine-Tuning:Training set}}       
	& & & & & & & & & & & & & & & & & & \\
	InternVL2-4B(before)[32f]                                               & 4 & 28.0 & 17.5 & 32.9 & 34.2 & 61.5 & 26.9 & 66.2 & 41.7 
		  & 21.0 & 25.3 & 37.0 & 37.3 & 33.3 & 25.8 & 30.8 & 35.0 & 21.4 \\
	InternVL2-4B(after)[32f]                                               &\cellcolor[HTML]{FE0000}{1} & 31.4 & 22.0 & 34.3 & 40.8 & 60.0 & 24.7 & 73.0 & 53.6 
		  & 21.0 & 44.1 & 51.1 & 31.3 & 42.9 &  34.4 & 36.5 & 35.0 & 19.6\\
	InternVL2-8B(before)[32f]                                               &\cellcolor[HTML]{FFCCC9}{3} & 29.4 & 22.4 & 25.7 & 35.5 & 53.8 & 22.6 & 59.5 & 48.8 
	& 23.2 & 30.0 & 30.4 & 38.8 & 40.5 & 33.3 & 35.9 & 34.2 & 20.8 \\
	InternVL2-8B(after)[32f]                                               &\cellcolor[HTML]{FD6864}{2} & 31.2 & 21.1 & 35.7 & 42.1 & 61.5 & 23.7 & 74.3 & 51.2 
	& 21.4 & 42.4 & 52.6 & 29.9 & 38.1 & 34.4 & 36.3 & 35.8 & 19.3 \\
	\rowcolor[HTML]{ECF4FF} 
	\multicolumn{1}{l|}{\cellcolor[HTML]{ECF4FF}\textit{Fine-Tuning:Test set}}       
	& & & & & & & & & & & & & & & & & & \\
	InternVL2-4B(before)[32f]                                               &\cellcolor[HTML]{FFCCC9}{3} & 28.3 & 21.3 & 45.2 & 31.8 & 63.0 & 20.4 & 66.7 & 43.8 
		  & 27.1 & 27.9 & 28.5 & 34.9 & 39.6 & 24.1 & 27.4 & 29.7 & 23.0 \\
	InternVL2-4B(after)[32f]                                               &\cellcolor[HTML]{FD6864}{2} & 31.5 & 25.5 & 38.1 & 34.1 & 60.9 & 20.4 & 66.7 & 37.5 
		  & 22.1 & 38.8 & 33.1 & 32.6 & 50.0 & 31.4 & 28.1 & 39.9 & 28.9 \\
	InternVL2-8B(before)[32f]                                               & 4 & 26.5 & 24.5 & 19.0 & 34.1 & 50.0 & 22.4 & 55.6 & 37.5  
	& 22.8 & 24.5 & 26.2 & 25.6 & 54.2 & 22.6 & 24.3 & 37.0 & 22.1 \\
	InternVL2-8B(after)[32f]                                               &\cellcolor[HTML]{FE0000}{1} & 31.7 & 25.5 & 35.7 & 34.1 & 60.9 & 18.4 & 66.7 & 39.6 
	& 23.8 & 37.4 & 31.5 & 34.9 & 50.0 & 32.8 & 27.7 & 39.1 & 29.4 \\
	\hline
		   \end{tabular}}
           }
           \vspace{-10pt}
	\end{table*}

We initially evaluated the performance of 16 popular Video-LLMs on various tasks related to embodied cognition in motion. Subsequently, we conducted detailed analyses focusing on the models, tasks, and video data sources. Finally, we summarized and categorized the reasons for failures across different tasks.

\subsection{Experimental Setup}

\textbf{Evaluation Metric:} Benefiting from the multiple-choice format of each question, we can directly calculate the accuracy for each task type as well as the overall accuracy.

\textbf{Baselines}: The baseline includes both proprietary and open-source Video-LLMs. For proprietary models, we used state-of-the-art models, including GPT-4o, GPT-4o-mini \cite{OpenAI_API}, Gemini-1.5 Flash, Gemini-1.5 Pro \cite{team2024gemini}, Gemini-2.0 Flash \cite{Gemini_API} and Qwen-VL-Max \cite{Qwen_Website}. For the open-source models, we focus on those capable of video/multiple image input, including LLaVA-NeXT-Video-7B \cite{liu2024llavanext}, Kangaroo \cite{liu2024kangaroo}, Qwen2-VL series~\cite{wang2024qwen2} and InternVL2 series \cite{chen2024far}.  
In terms of input frame numbers, the Gemini and Qwen series allow for convenient adjustment of the input parameter \textit{fps}, while others require input as \textit{frame number}. To ensure a fair comparison while considering local computational resource constraints, the goal is to input as many frames as possible.

\textbf{Others}: All local model inference and fine-tuning is performed on three NVIDIA RTX A6000.
Detail calculation of evaluation metric, baseline settings, and prompts can be found in Appendix \ref{appendix:exp_setting}.

\subsection{Model Comparison}
The quantitative result is shown in Table \ref{tab:example_table_large_font}. We can draw the following conclusions:
\begin{itemize}[leftmargin=*]
	\item Both proprietary models and open-source models exhibit relatively poor embodied cognitive abilities when navigating urban open-ended spaces. The best-performing model, Qwen-VL-Max, achieves an average accuracy of only 45.5\%. This underscores the value of UrbanVideo-Bench, highlighting that embodied cognitive abilities in urban three-dimensional spaces have not been adequately addressed.
	\item Some open-source Video-LLMs outperform part of proprietary models. Specifically, models that have been optimized for video data demonstrate superior performance compared to LMMs that focus on images.
    \item Smaller parameter models appear to be more unstable. For two models with equivalent average accuracy, the open-source small parameter model tends to have a lower minimum accuracy across all tasks compared to the commercial large parameter model.
	
\end{itemize}

\subsection{Correlation of Cognitive Abilities}
\begin{figure}[t]
	\centering
	\includegraphics[width = \linewidth]{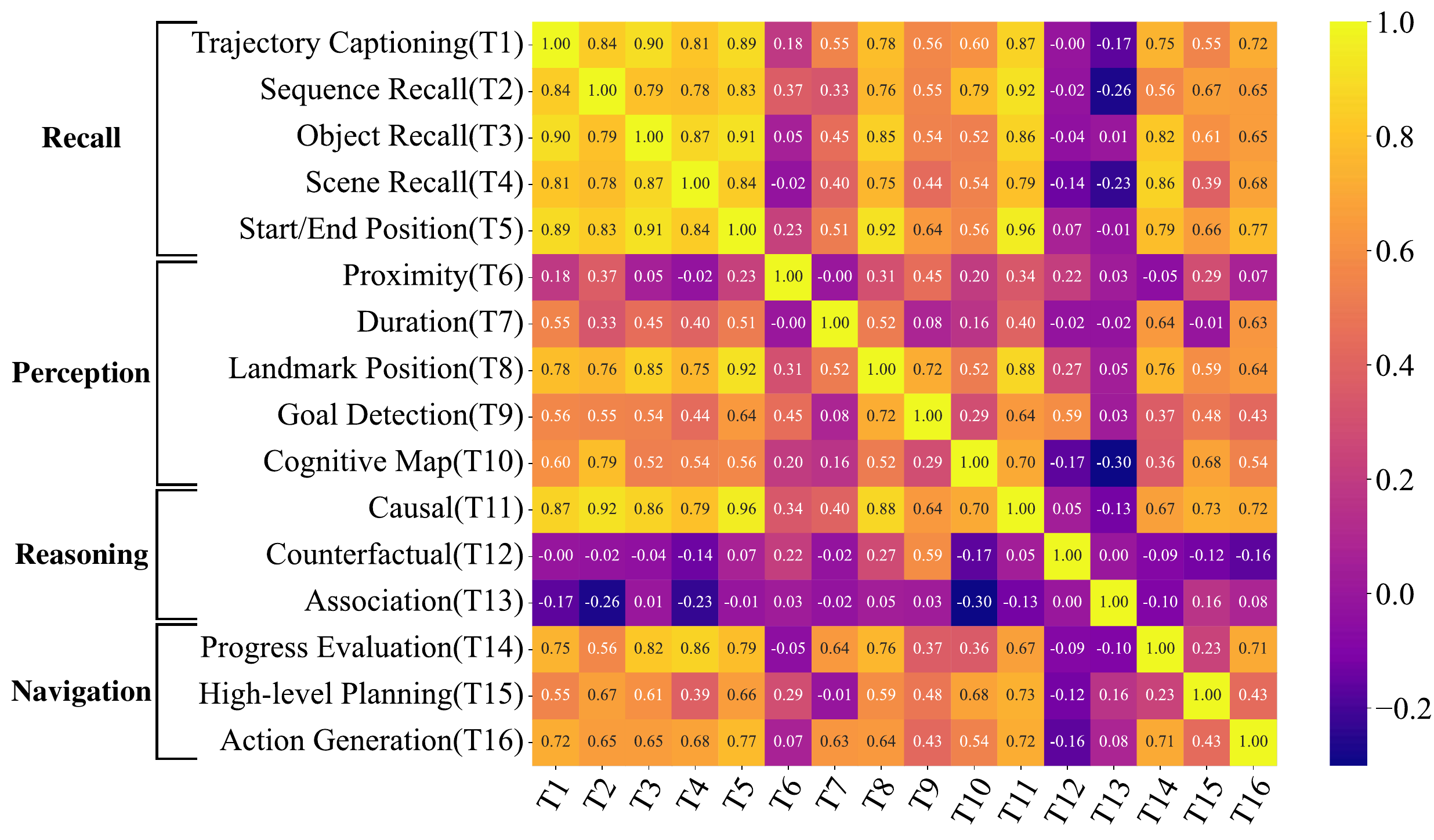}
	\caption{Correlation of Cognitive Abilities. A higher value indicates more similar Video-LLMs' performance on the two tasks, implicitly indicating the relationship between the two tasks.}
	\label{Fig:experiment}
\end{figure}
We tend to explore the relationships between these tasks and the underlying cognitive abilities, as shown in Figure \ref{Fig:experiment}. Specifically, we compute the pairwise correlations between each column (representing individual tasks) in Table \ref{tab:example_table_large_font}, excluding the fine-tuning portion. The implicit assumption in this approach is that if multiple models exhibit similar performance across two tasks, it suggests that the embodied cognitive abilities needed for these tasks are similar. We can derive the following conclusions:
\begin{itemize}[leftmargin=*]
	\item The \textbf{causal reasoning} task exhibits a high correlation with almost all other tasks. It suggests that the \textbf{ability to understand and infer causality is fundamental to a wide range of cognitive processes}. This finding may indicate that causal reasoning is potentially a key factor in the emergence of embodied cognitive in Motion.
	\item \textbf{Recall}-type tasks demonstrate strong inter-correlations among themselves. These tasks all involve memory-related problems, underscoring the ability to recall information is a shared underlying requirement for these tasks, highlighting memory as a central component in cognitive task execution.
        \item \textbf{Navigation} tasks have a high correlation with both \textbf{Recall and Perception} tasks. This observation aligns with prior knowledge that effective action and planning depend on robust memory and perceptual capabilities. 
        \item \textbf{Counterfactual and Association} reasoning —both high-level reasoning tasks—exhibit low correlations with other task types. These tasks rely on distinct cognitive processes that are not shared with the other tasks in our analysis. This suggests that some embodied cognitive abilities may operate independently rather than as components of a general intelligence framework. Therefore, \textbf{when tasks involve these two high-level abilities, targeted training is necessary.}
	
\end{itemize}

\subsection{Sim-to-Real}
\begin{figure}[t]
	\centering
	\includegraphics[width = \linewidth]{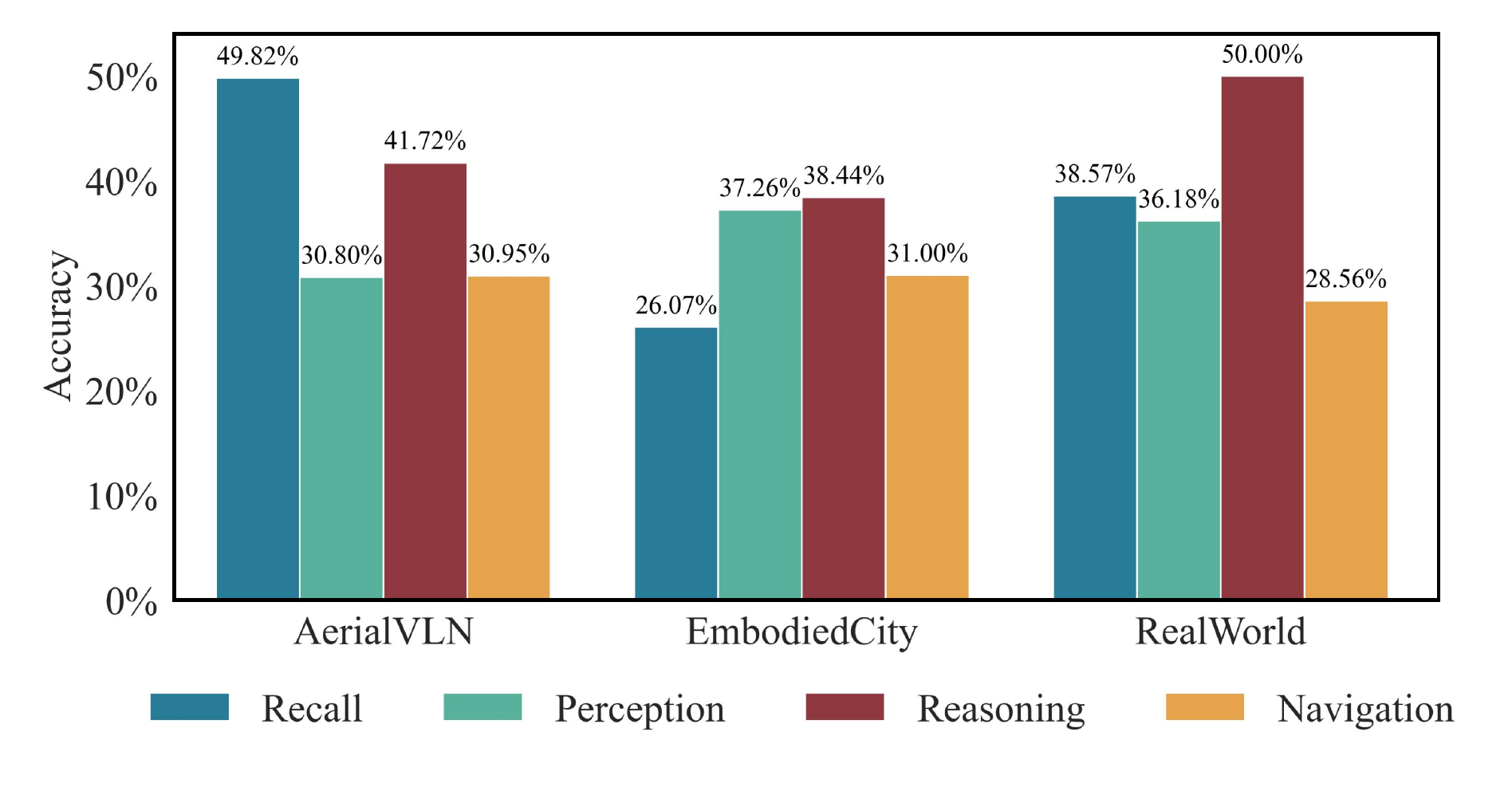}
	\caption{Average performance of 17 Video-LLMs in different video sources. }
	\label{Fig:data_source}
\end{figure}
We average the performance of all Video-LLMs and listed the results of four categories of cognitive abilities across three different data sources, as shown in Figure \ref{Fig:data_source}. From the model performance perspective, there is no significant distributional difference among the various video sources overall. Embodied research has long suffered from a lack of real-world data. Here, we used data from EmbodiedCity and AerialVLN as the training set, and real-world data as the test set. We employed LoRA~\cite{hu2021lora} to fine-tune the large models InternVL-4B and InternVL-8B to explore the potential for Sim2Real transfer (see Appendix \ref{appendix:Fine-Tuning} for details on fine-tuning.). The results, as shown in Table \ref{tab:example_table_large_font}, indicate that both Goal Detection and Association Reasoning on the test set improved by approximately 7\% post-fine-tuning. The mean improvements for the two fine-tuned models were 3.2\% and 5.2\%, respectively.

\subsection{Error Analysis}
\begin{figure}[t]
	\centering
	\includegraphics[width =\linewidth]{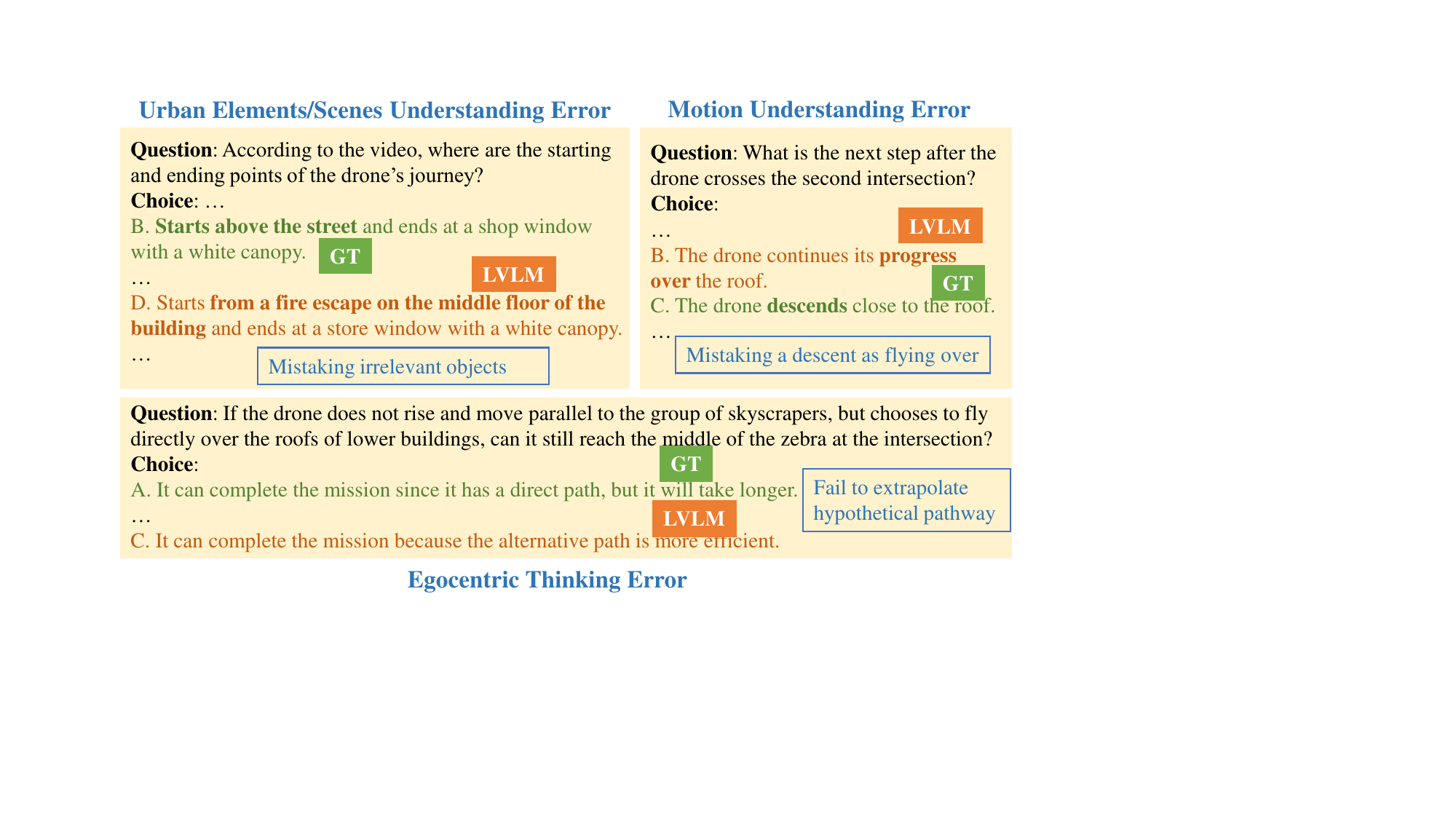}
	\caption{Three common errors in Video-LLMs when performing tasks of urban embodied cognition, along with the corresponding examples.}
	\label{Fig:error_analysis}
\end{figure}
After examining the reasoning processes of Video-LLMs, three common error types were identified, as shown in Figure \ref{Fig:error_analysis}:

\begin{itemize}[leftmargin=*]
	\item \textbf{Urban Elements/Scenes Understanding Error}: Complex urban scenes pose challenges to perception-related tasks for Video-LLMs, resulting in insufficient alignment and urban hallucinations. This means that the models guess based solely on textual content and fail to detect urban objects that are entirely absent in the video during analysis.
	\item \textbf{Motion Understanding Errors}: Video-LLMs struggle to distinguish orientation and misinterpret changes in the camera's gimbal angle as vertical movement, indicating limited spatial awareness.
        \item \textbf{Egocentric Thinking Error}: Video-LLMs fail to perform complex embodied reasoning tasks, such as route planning and extrapolation.
	
\end{itemize}

Recall and Perception tasks, reliant on visual abilities, suffer from urban elements/scenes and agent motion understanding error. As for the complex and challenging tasks of reasoning and navigation, various errors are prevalent.
See Appendix \ref{appendix:Error_Cases} for more details.

%% file: 5.Conclusion.tex
\section{Conclusion}

In this work, we propose UrbanVideo-Bench, a benchmark for embodied motion cognition in urban open spaces, comprising 1.5k video clips and 5.2k multiple-choice questions. We evaluated the performance of 17 currently popular Video-LLMs in terms of recall, perception, reasoning, and planning. The experimental results indicate that the best current Video-LLMs achieve only a 45.5\% accuracy rate. Our analysis further reveals that causal reasoning is highly correlated with other tasks such as recall, perception, and planning. Fine-tuning large models with simulation data can enhance their performance on real-world embodied video tasks.

%% file: Appendix.tex
\section{Dataset Examples} \label{appendix:Dataset_Examples}

\begin{figure*}[t]
\centering
\includegraphics[width=\linewidth]{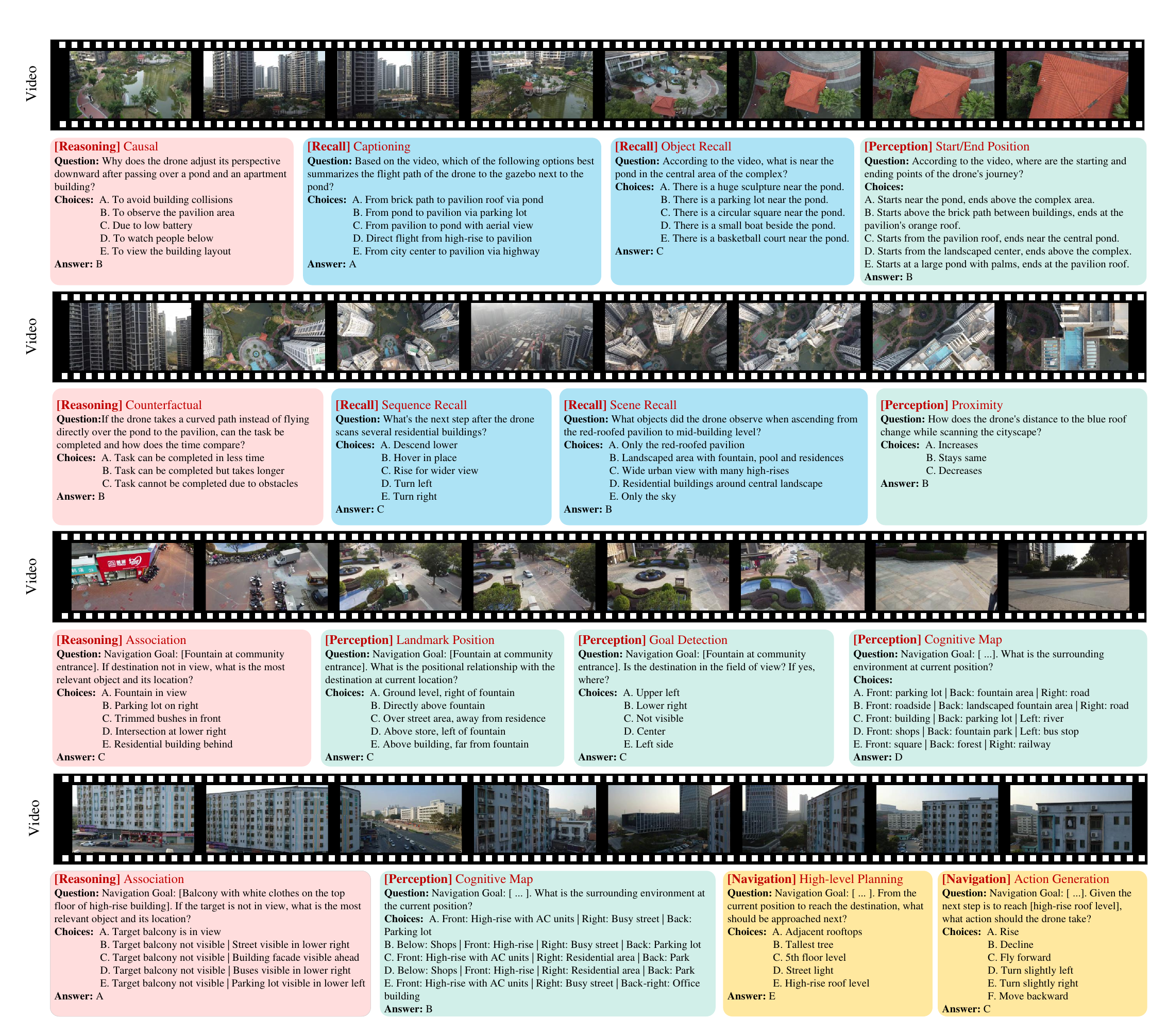}
\caption{Videos collected from the real drones and MCQs examples.}
\label{fig:real_world_examples}
\end{figure*}

\begin{figure*}[t]
\centering
\includegraphics[width=\linewidth]{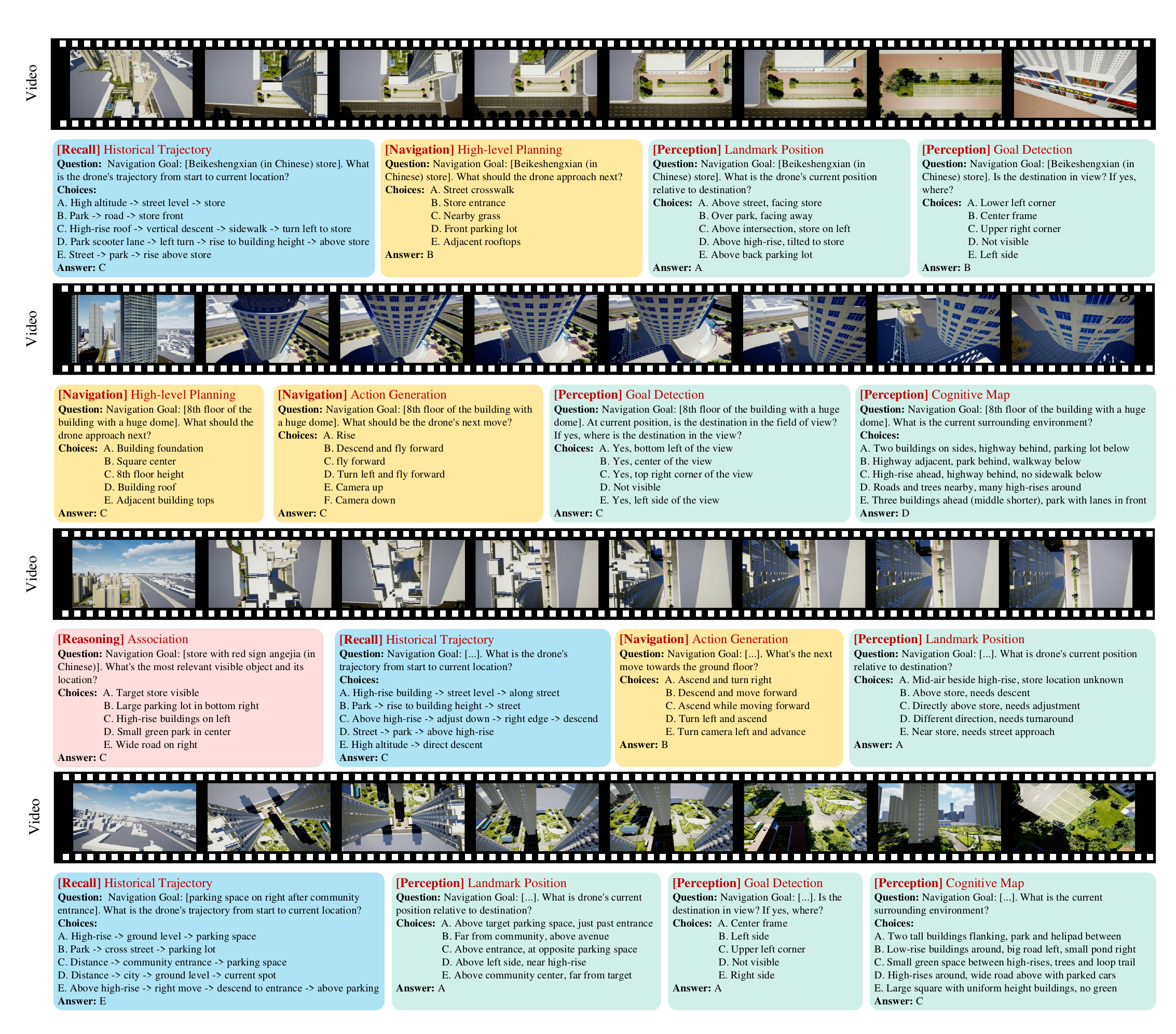}
\caption{Videos collected from the EmbodiedCity simulator and MCQs examples.}
\label{fig:embodiedcity_examples}
\end{figure*}

\begin{figure*}[t]
\centering
\includegraphics[width=\linewidth]{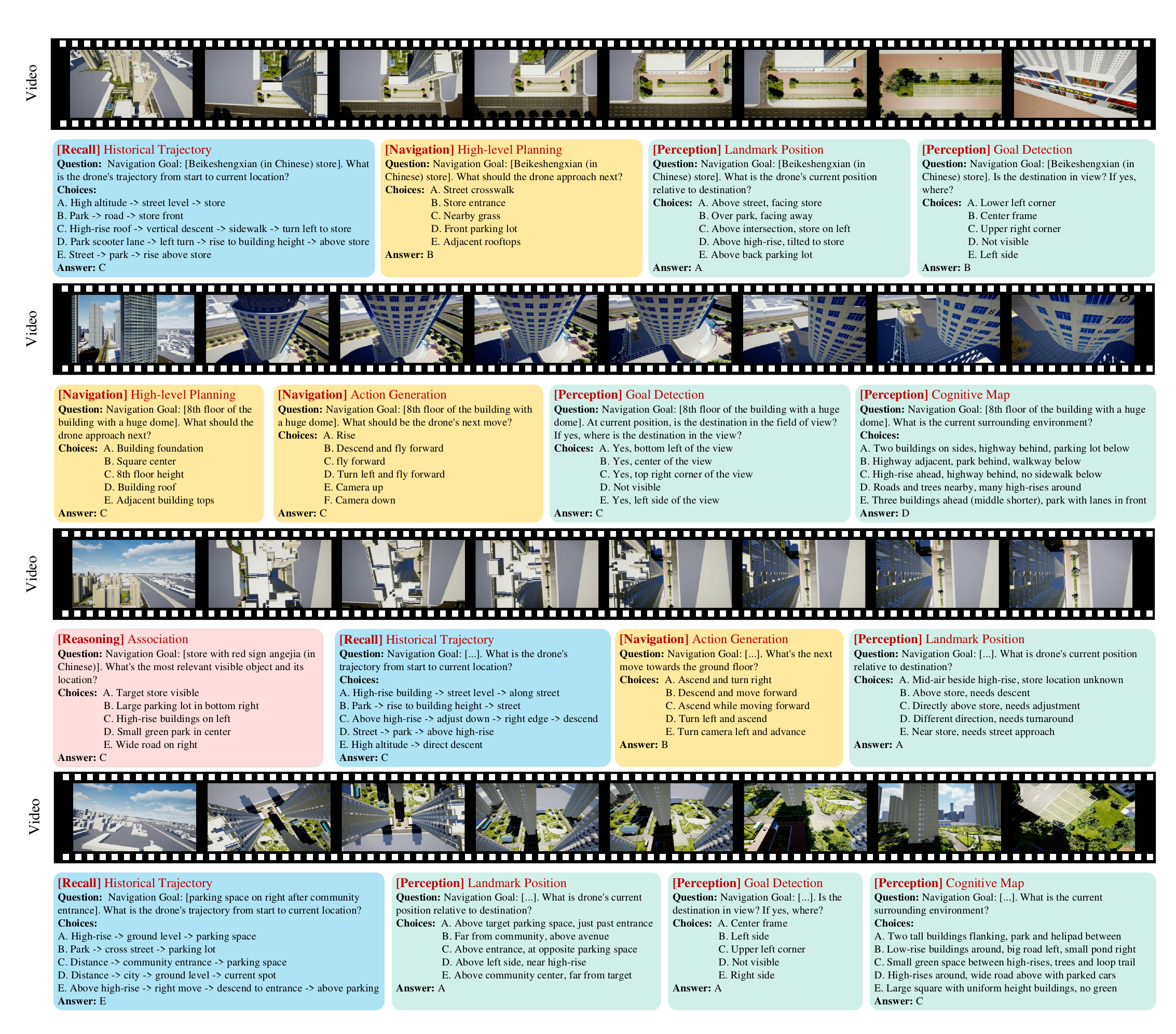}
\caption{Videos collected from the AerialVLN simulator and MCQs examples.}
\label{fig:aerialvln_examples}
\end{figure*}

To better illustrate the proposed dataset, we provide MCQ examples and videos from real world, the EmbodiedCity simulator, and the AerialVLN simulator, encompassing all task types. They are presented in Fig. \ref{fig:real_world_examples}, Fig. \ref{fig:embodiedcity_examples}, and Fig. \ref{fig:aerialvln_examples}.

\section{Dataset Generation Pipeline} \label{appendix:pipeline}

\subsection{Details of Video Curation} \label{appendix:video_curation}

\textbf{Selected Simulator Advantages:}
a) Realistic environment modeling. Both EmbodiedCity and AerialVLN are built on Unreal Engine and include various building types and streets, with more than a hundred categories of micro urban elements, which enriches the semantic information of the obtained embodied mobility video data.
b) Support for aerial agents. Both simulators have built-in AirSim plugins, allowing easy control of aerial agents.
c) Existing route reference. Research work on vision-language navigation \cite{zhu2020vision} has already been conducted in these simulators, allowing us to obtain some route coordinates and instruction data. Although most routes cannot be directly used (for example, most flying paths in AerialVLN contain many meaningless, repetitive flying actions and lack logicality), they provide some reference for our data collection.

\textbf{Drone Setup:} The drone has only one camera equipped with a gimbal, which can tilt from 0 degrees to 90 degrees downward. The second group consists of experienced pilots (with over 1000 hours of real drone flight time) to ensure the rationality of flight operations. Through this approach, the movement of the agent is purposeful, and the collected video data is more conducive to evaluating the capabilities of Video-LLMs.

\subsection{Details of MCQ Generation} \label{appendix:MCQ_generation}
In this work, we use LLM to generate the multiple choice questions (MCQ) in the dataset. 

a) Narration Generation: Initially, the LMM is prompted to systematically generate a narration of the UAV's trajectory by combining embodied movement videos and destination instructions, based on the videos from Embodied City. Videos from the other two sources are originally combined with trajectory infomation.

b) Structured Extraction of Key Information: The LMM then extracts a list of movements and objects, providing structured text that ensures the subsequent MCQ generation is more aligned with the requirements.

c) Role-playing: In the last MCQ generation prompt, the model is given a specific context and role, enhancing its understanding of the task and adherence to instructions.

d) Providing Question/Option Templates and Examples: For each task, several templates for questions and options are provided, with sections marked for replacement. We also provide detailed task definitions and examples. 

\begin{figure*}[t]
\centering
\includegraphics[width=\linewidth, trim=0 380pt 0 0, clip]{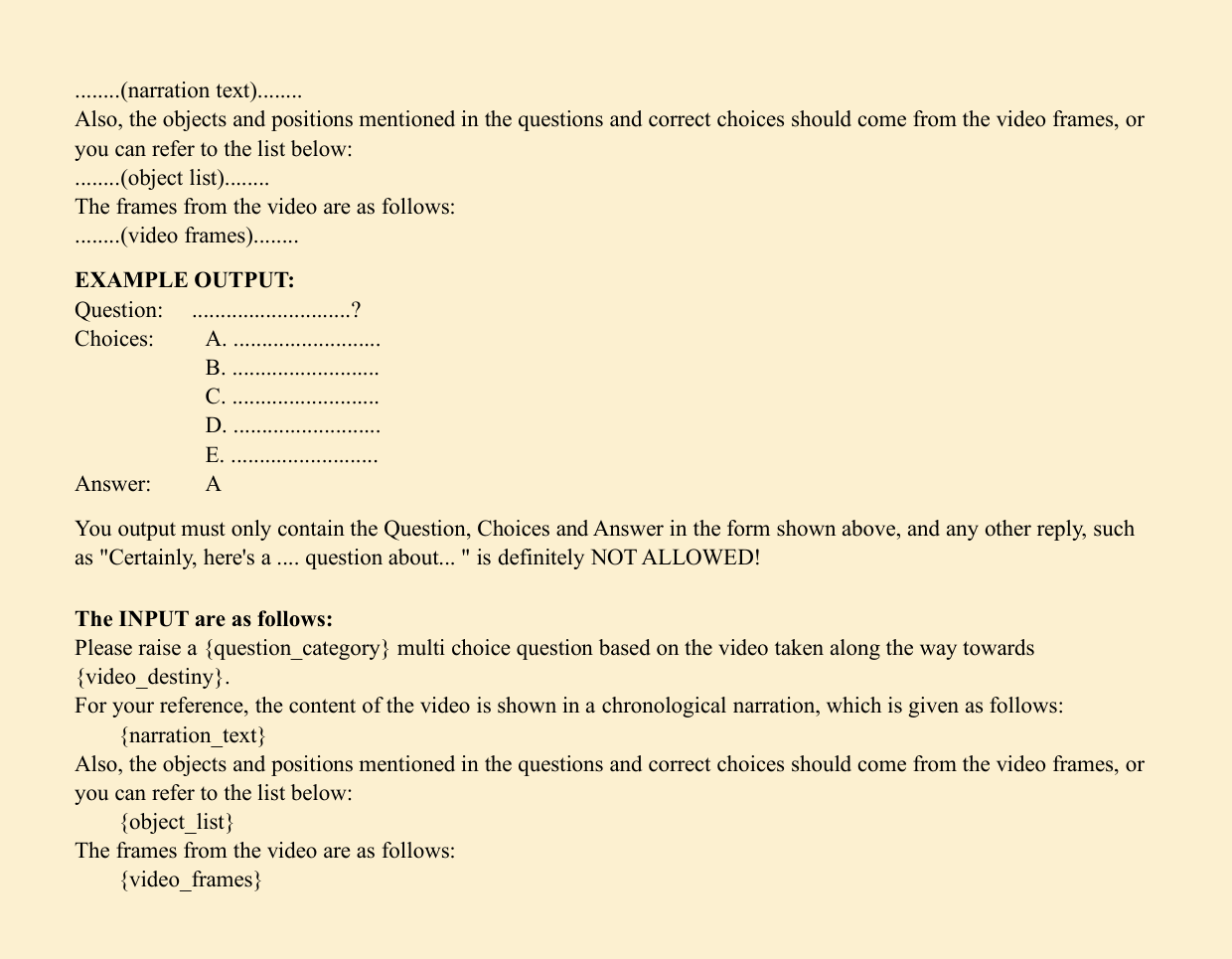}
\caption{The prompt used in MCQ generation. The prompt contains scenario setup, detailed instructions, respectively written template and example output, and finally input template. Role playing and formatted structure help the LLM to better understand the user’s intentions.}
\label{fig:question generation prompt}
\end{figure*}

In the prompt, we first set up a scenario for the model, one where the model act as a teacher who needs to raise a series of test questions based on the given videos. This role playing trick makes it easier for us to explain the details of the question generation task onward. Then, we break the task into several sequential parts, and give the specific requirements for the sub-tasks. The sub-tasks include video understanding, question generation, answer generation, and finally structured input/output, with requirements, templates and examples. As we have 14 categories of questions focusing on different aspect of the video, it’s hard to cover them all in one general instruction, so we adopt respectively written instructions for question and answer generation, each having its own task explanation, template and example output. During tests, the LLM sometimes gives random explanatory text before it raises the question, which could be an obstacle in the processing work, so we added specific instructions in the prompt to prevent the model from doing so. The main part of the prompt we use are shown in Figure \ref{fig:question generation prompt}, though most of the explanations, templates and examples for question generating are left out due to space constraints.

\begin{figure*}[htbp]
\centering
\includegraphics[width=\linewidth]{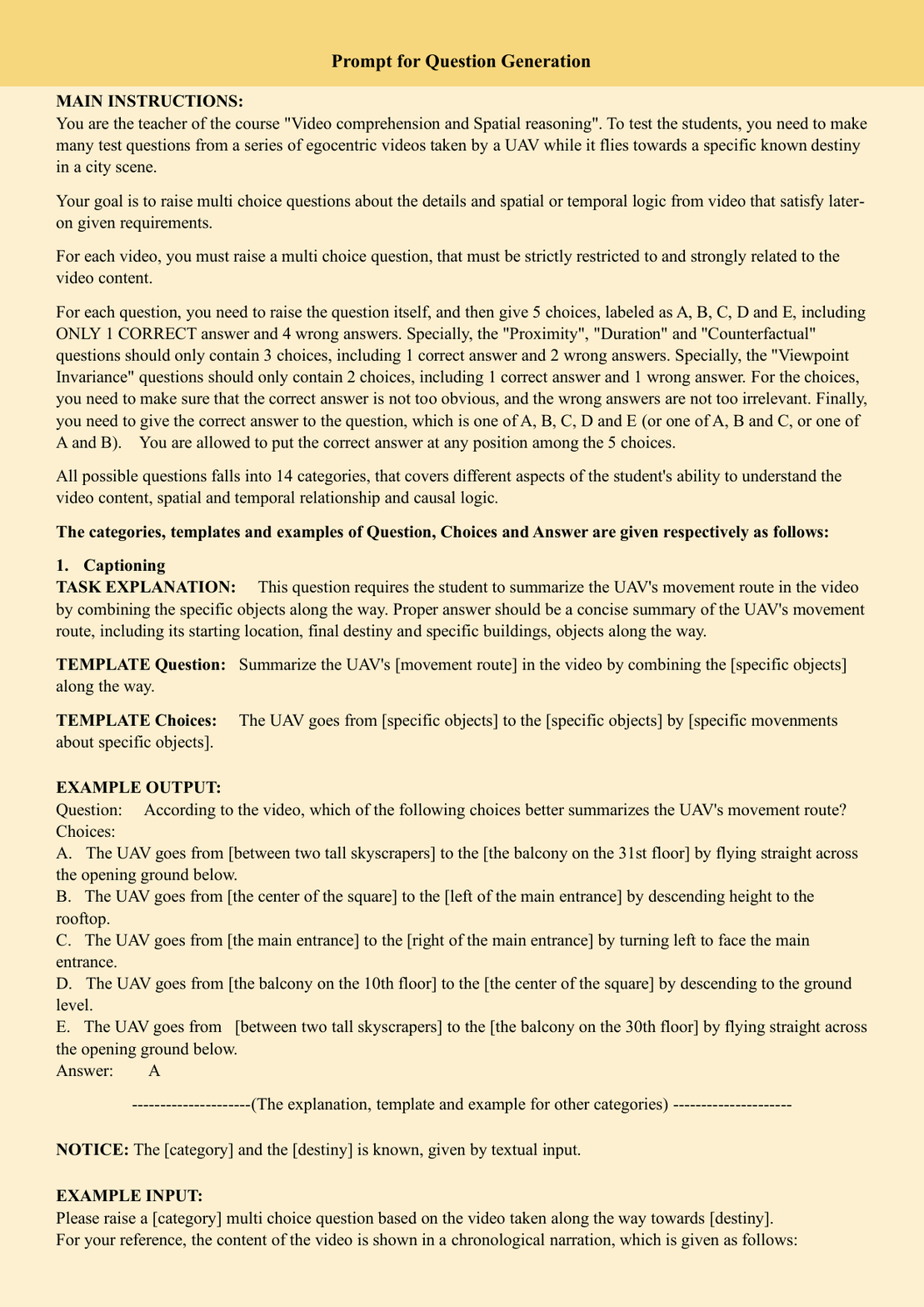}
\end{figure*}

\begin{figure*}[t]
\centering
\includegraphics[width=\linewidth, trim=0 350pt 0 0, clip]{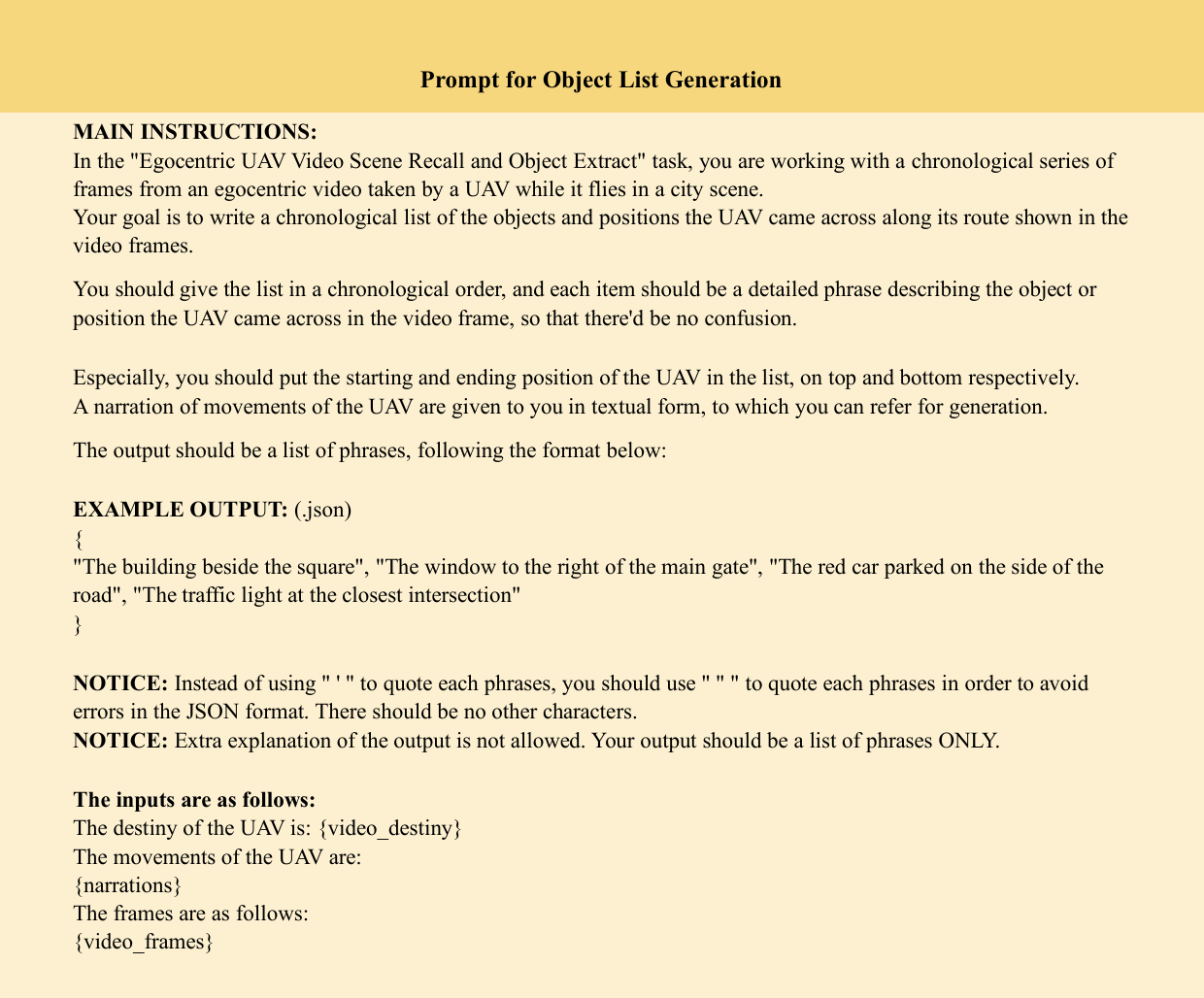}
\caption{The prompt used in object list generation. The prompt contains scenario setup, detailed instructions and example output. As a simpler task than question generation, input/output template is not needed.}
\label{fig:object prompt}
\end{figure*}

\begin{figure*}[t]
\centering
\includegraphics[width=\linewidth, trim=0 280pt 0 0, clip]{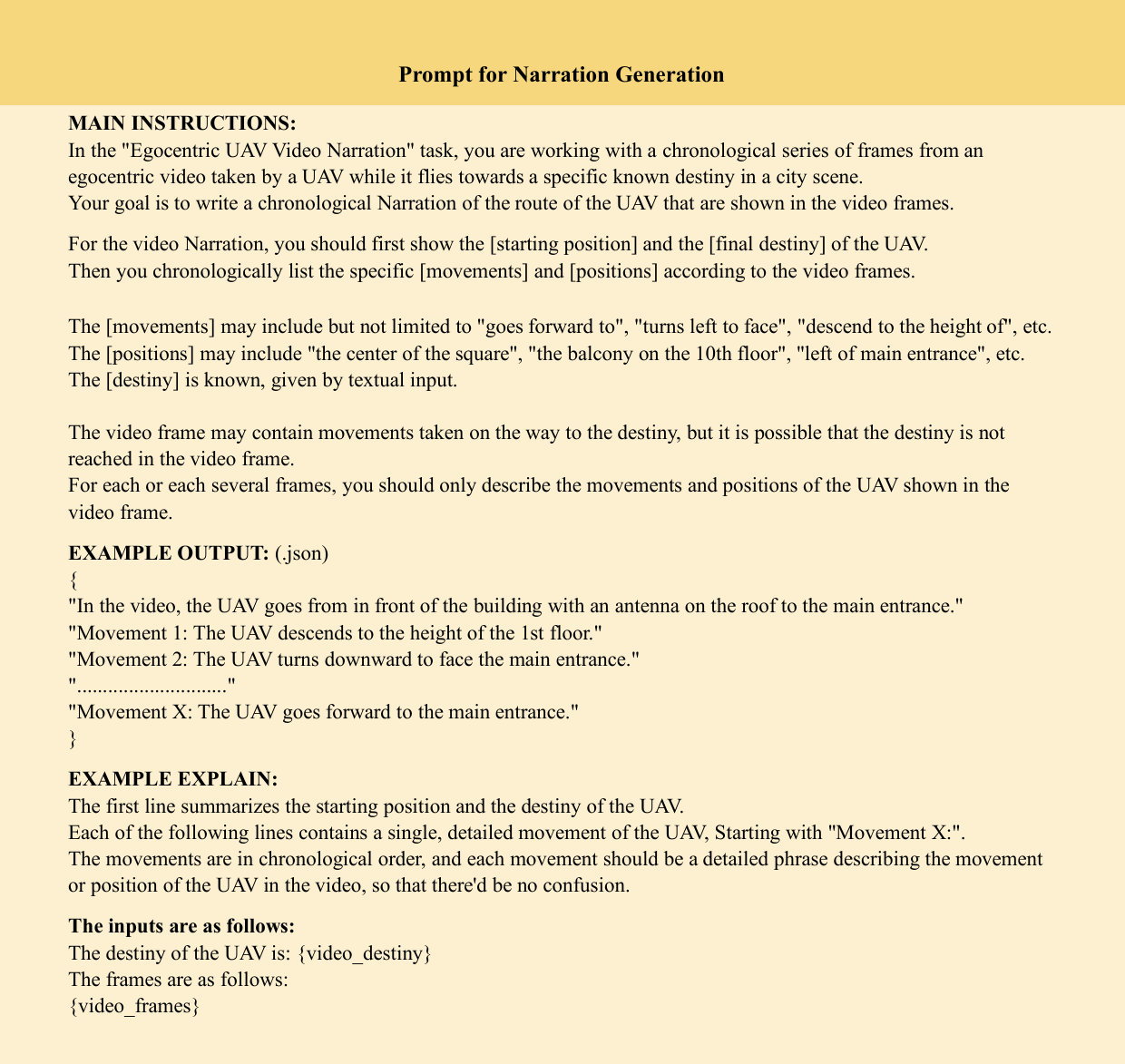}
\caption{The prompt used in movement extraction. The prompt contains scenario setup, detailed instructions, respective explanation of prompt components, and finally input template.}
\label{fig:movement prompt}
\end{figure*}

In question generation, we find that the LLM still has problem understanding complicated city environments in the videos, so we introduced pre-generated structured object list and movement series to assist the model in understanding the videos. Also, this step is done by prompts shown in Figure \ref{fig:object prompt} and Figure \ref{fig:movement prompt}.

\subsection{Details of Human Refinement} \label{appendix:human_refinement}
Following the automatic question generation phase, we implemented a systematic human refinement process to ensure the quality and reliability of the generated questions.  During the refinement phase, we identified and addressed various issues across the generated questions, with statistics shown in Table~\ref{tab:issues_stats}. This process focused on four major aspects requiring manual intervention, with detailed examples provided in Table~\ref{tab:refinement_examples}. First, ambiguous or unclear questions were identified and refined to ensure precise communication of the intended query. Second, hallucinated urban elements, which were occasionally generated by the LLM but not present in the actual video content, were either corrected or removed to maintain factual accuracy. Third, directional descriptions were standardized by converting absolute directions to relative ones, consistently using the drone as the reference system. This standardization was crucial for maintaining consistency in spatial relationships across the dataset. Finally, multiple-choice options were carefully reviewed and adjusted to ensure appropriate differentiation between choices while maintaining one unambiguously correct answer.

The refinement process significantly enhanced the dataset's quality by eliminating potential sources of confusion and ensuring all questions accurately reflected the video content. Human annotators were specifically trained to identify these issues and apply standardized corrections, resulting in a more robust and reliable benchmark for evaluating video-based navigation understanding.

\begin{table}[h]
	\caption{Statistics of Issues in Human Refinement Phase}
	\small
	\setlength\tabcolsep{4pt}
	\begin{tabular}{lr}
	\toprule
	Issue & Count \\
	\midrule
	Invalid/ambiguous questions & 761 \\
	Urban element hallucination & 184 \\
	Direction & 446 \\
	Choices with insufficient differentiation or errors & 2363 \\
	\bottomrule
	\end{tabular}
	\label{tab:issues_stats}
\end{table}

\begin{table*}[h]
	\caption{Examples of human refinement for generated questions.}
	\label{tab:refinement_examples}
	\small
\setlength\tabcolsep{4pt}
\begin{adjustbox}{width=\textwidth}
	\begin{tabular}{|c|c|l|}
	\rowcolor[HTML]{4874CB} 
	\hline
	{\color[HTML]{FFFFFF} \textbf{Issue}}                   & {\color[HTML]{FFFFFF} \textbf{Refinement}}            & \multicolumn{1}{c}{\cellcolor[HTML]{4874CB}{\color[HTML]{FFFFFF} \textbf{Example}}}\\ \hline
	\multicolumn{1}{|c|}{}                                  & \multicolumn{1}{c|}{}                                 & \multicolumn{1}{l|}{\begin{tabular}[c]{@{}l@{}}Navigation Goal: [8th floor of the building ({\color{red}with a huge dome})], what is the \\ \st{direction} ({\color{red}spatial relationship}) between the current position and the destination\\  when the drone reaches the current position?\end{tabular}}\\ \cline{3-3} 
	\multicolumn{1}{|c|}{\multirow{-4}{*}{Invalid/ambiguous questions}}                                             & \multicolumn{1}{c|}{\multirow{-4}{*}{\begin{tabular}[c]{@{}c@{}}Clarify and complete unclear or \\ ambiguous parts in question stems\end{tabular}}} 
	& \multicolumn{1}{l|}{\begin{tabular}[c]{@{}l@{}}Navigation Goal: [Rooftop of a nearby blue building]. Given the [\st{planned milestone} \\ ({\color{red}top of the tallest building ahead})], what should be the drone's next action?\end{tabular}}\\ \hline
	
	\multicolumn{1}{|c|}{}                                  & \multicolumn{1}{c|}{\begin{tabular}[c]{@{}c@{}}Correct hallucinated elements \\ to accurate elements\end{tabular}}                                  & \multicolumn{1}{l|}{\begin{tabular}[c]{@{}l@{}}Navigation Goal: [A store with a green sign reading "\st{Meiyue Stylin}g" ("{\color{red}Orange Fresh}")]\\ What should the drone approach next?\end{tabular}}\\ \cline{2-3} 
	\multicolumn{1}{|c|}{\multirow{-3}{*}{Urban element hallucination}}                                         & \multicolumn{1}{c|}{Remove hallucinated elements}     & \multicolumn{1}{l|}{\begin{tabular}[c]{@{}l@{}}A. Two buildings on sides, highway behind, parking lot below\\ B. Highway adjacent, park behind, walkway below\\ C. High-rise ahead, highway behind, no sidewalk below\\ D. Roads and trees nearby, many high-rises around,\st{parking lot below}.\\ E. Three buildings ahead , park with lanes in front\\ Answer: D\end{tabular}}\\ \hline
	
	\multicolumn{1}{|c|}{}                                  & \multicolumn{1}{c|}{}                                 & \multicolumn{1}{l|}{\begin{tabular}[c]{@{}l@{}}\st{The drone is positioned on the right side of the highway} \\ ({\color{red}The highway is positioned on the right side of the drone}).\end{tabular}}\\ \cline{3-3} 
	\multicolumn{1}{|c|}{}                                  & \multicolumn{1}{c|}{\multirow{-3}{*}{\begin{tabular}[c]{@{}c@{}}Use drone as reference system \\ for directions\end{tabular}}}                      
	& \multicolumn{1}{l|}{\begin{tabular}[c]{@{}l@{}}\st{The target building is positioned on the left side of the drone} \\ ({\color{red}drone has the target building in its left field of view}).\end{tabular}}\\ \cline{2-3} 
	\multicolumn{1}{|c|}{\multirow{-5}{*}{Direction}}       & \multicolumn{1}{c|}{\begin{tabular}[c]{@{}c@{}}Convert absolute directions \\ to relative directions\end{tabular}}                                & \multicolumn{1}{l|}{\begin{tabular}[c]{@{}l@{}}\st{The drone is currently located slightly southwest of the target balcony} \\ ({\color{red}The target building is positioned on the lower left side of the drone}).\end{tabular}}\\\hline
	
	\multicolumn{1}{|c|}{}                                  & \multicolumn{1}{c|}{}                                 & \multicolumn{1}{l|}{\begin{tabular}[c]{@{}l@{}}Choices:  \\ A. park walkway -> high-rise\\ B. Park -> street -> high-rise level -> building side\\ C. Street -> park -> high-rise level -> 15th floor balcony  \\ D. \st{Park -> street -> high-rise level -> roof edge -> 15th floor balcony} \\ ({\color{red}Left side of domed building -> right turn-> slight descent -> horizontal move to building})\\ E. High-rise -> descent -> park\\ Answer: D\end{tabular}} \\ \cline{3-3} 
	\multicolumn{1}{|c|}{}                                  & \multicolumn{1}{c|}{\multirow{-8}{*}{\begin{tabular}[c]{@{}c@{}}Correct incorrect options \\ to be accurate\end{tabular}}}                          & \multicolumn{1}{l|}{\begin{tabular}[c]{@{}l@{}}Choices:  \\ A. Grass ascent is longer\\  B. Tree area crossing is longer \\  C. Equal duration\\ Answer: \st{A}({\color{red}B})\end{tabular}}\\ \cline{2-3} 
	\multicolumn{1}{|c|}{\multirow{-14}{*}{\begin{tabular}[c]{@{}c@{}}Choices with insufficient\\  differentiation or errors\end{tabular}}} & \multicolumn{1}{c|}{\begin{tabular}[c]{@{}c@{}}Modify one option \\ to contain factual errors\end{tabular}}   & \multicolumn{1}{l|}{\begin{tabular}[c]{@{}l@{}}Choices:  \\A. Ascend\\ B. ({\color{red}Backward after}) descend \\ C. Forward ({\color{red}after descent})\\ D. Left turn \\ E. Right turn\\ F. Rotate camera upward\\ G. Rotate camera downward\\ Answer: C\end{tabular}}\\ \hline
	\end{tabular}
	\end{adjustbox}
\end{table*}

\section{Additional Experimental Settings} \label{appendix:exp_setting}

\subsection{Evaluation Metric Calculation}

This study employs a multi-stage workflow to evaluate the accuracy of a model in video-based multiple-choice question answering tasks. This workflow integrates regular expressions and model inference for efficiency, complemented by manual validation to ensure robustness. The combined approach achieves precise and scalable accuracy assessment for the target task.The procedure is structured as follows:

\begin{itemize}[leftmargin=*]%
\item \textbf{Answer Extraction and Correction.} First, answers are extracted from the model-generated text. Regular expressions are applied to identify standardized outputs (e.g., options starting with ``A.'' or standalone alphabetic characters). For unstructured or ambiguous text, the GPT-4o model is invoked to infer the most plausible option (A--G) based on the output context. Entries unresolved by automated methods are flagged with the special token ``l**k''. A manual verification step is subsequently performed to correct these flagged entries, ensuring the validity of all answers.

\item \textbf{Data Alignment and Cleaning.} The corrected answers are aligned with a reference dataset containing ground-truth answers. Data consistency is ensured through format standardization (e.g., removing ``.mp4'' suffixes from video IDs and normalizing column names). Valid entries are filtered using an inner join operation, merging model predictions with reference data based on video IDs and question categories.
    
\item \textbf{Statistical Analysis and Result Export.} Accuracy is computed by category through group-wise comparisons of matched answers, followed by aggregation to derive the overall accuracy. Results are organized into structured tables, detailing per-category and total accuracy values. These outputs facilitate systematic performance evaluation and further analysis.
\end{itemize}

\subsection{Brief Introduction on Baselines}

We introduce the participating long-context LMMs as follows:

\textbf{Gemini-1.5-Flash}. Released on February 14, 2024, with an API service, Gemini-1.5-Flash is a model with 150 million parameters. It supports an input token limit of 1 million, an output token limit of 8192, and a maximum video duration of 1 hour. The frame rate is set to 1 fps.

\textbf{Gemini-1.5-Pro}. Released on February 14, 2024, with an API service, Gemini-1.5-Pro is a model with 175B parameters. It supports an input token limit of 2 million, an output token limit of 8192, and a maximum video duration of 2 hours. The frame rate is set to 1 fps.

\textbf{Gemini-2.0-Flash}. Released on December 11, 2024, Gemini-2.0-Flash is the latest lightweight model in the Gemini series, offering improved efficiency and a 12M context length. The input token limit is 1 million, the output token limit is 8192, and the frame rate is 1 fps.

\textbf{GPT-4o-mini}. Released on July 18, 2024, GPT-4o-mini is a compact version of GPT-4o, designed for faster inference with a 64K context length. It has 8B parameters, an input token limit of 128K, an output token limit of 16384, and a frame rate of 32 f.

\textbf{GPT-4o}. Released on May 13, 2024, with an API service, GPT-4o is the latest multimodal LMM from OpenAI, featuring a 128K context length. It has 200B parameters, an input token limit of 128K, an output token limit of 4096, and a frame rate of 32 f.

\textbf{Qwen-VL-Max-latest}. Released in April 2024, Qwen-VL-Max-latest is the most advanced model in the Qwen-VL series, supporting multimodal tasks with a 128K context length. It has an input token limit of 128K, an output token limit of 8192, and a frame rate of 32 f.

\textbf{LLaVA-NeXT-Video-7B-hf}. Released in April 2024, LLaVA-NeXT-Video-7B-hf is a video-focused LMM with 7B parameters. In this experiment, we set the frame rate to 32 f and the output token limit to 512.

\textbf{Phi-3.5-vision-instruct}. Released in August 2024, Phi-3.5-vision-instruct is an upgraded version of Phi-3-Vision-Instruct, with 4.2B parameters and a 128K context length. In this experiment, we set the frame rate to 32 f and the output token limit to 512. The temperature coefficient is set to 0, and the maximum possible output is selected.

\textbf{Kangaroo}. Released on July 17, 2024, Kangaroo is a specialized LMM for multi-image and video understanding, with 8B parameters. Since the frame rate cannot be modified in the code, in this experiment, we used the default frame rate of 64 f and set the output token limit to 256. The temperature coefficient is set to 0, and the maximum possible output is selected.

\textbf{Qwen2-VL-2B-Instruct}. Released on August 30, 2025, Qwen2-VL-2B-Instruct is a lightweight instruct-tuned model with 2B parameters and a 64K context length. In this experiment, we set the resolution to 360 * 420 and adopted the model's default settings. The frame rate is 0.5 fps, which is the maximum limit for our GPU's computational capacity. The output token limit is set to 256.

\textbf{Qwen2-VL-7B-Instruct}. Released on August 30, 2025, Qwen2-VL-7B-Instruct is a mid-sized instruct-tuned model with 7B parameters and a 128K context length. In this experiment, we set the resolution to 360 * 420 and adopted the model's default settings. The frame rate is 0.25 fps, which is the maximum limit for our GPU's computational capacity. The output token limit is set to 256.

\textbf{InternVL2 series (2B, 4B, 8B, 26B, 40B, Llama3-76B)}. Released on July 4, 2025, the InternVL2 series offers a range of models from 2B to 76B parameters, with context lengths scaling from 32K to 256K, designed for diverse multimodal tasks. In this experiment, we set the resolution to 448 * 448 and adopted the model's default settings. The frame rate is 32 f, and the output token limit is set to 1024.

\subsection{Responses of Video-LLMs}
Table \ref{tab:Format_Correctness} presents the format correctness rates of six proprietary models through API calls. The probabilities indicate each model's capability to reliably produce outputs adhering to the specified format requirements. These results demonstrate the varying levels of format adherence among state-of-the-art language models.
\begin{table}[h]
        \centering
	\caption{Format Correctness Rates of Proprietary Models}
	\small
    \scalebox{0.92}{
	\begin{tabular}{lr}
	\hline
	\textbf{Proprietary Models (API)} & \textbf{Rate} \\
	\midrule
	Gemini-1.5-Flash       & 0.992762 \\
        Gemini-1.5-Pro         & 0.983810 \\
        Gemini-2.0-Flash       & 0.969714 \\
        GPT-4o-mini            & 0.912190 \\
        GPT-4o                 & 0.961143 \\
        Qwen-VL-Max-latest     & 0.997333 \\
	\hline
	\end{tabular}
    }
	\label{tab:Format_Correctness}
\end{table}

\subsection{Cost of Proprietary Models}
In Table \ref{tab:evaluation_cost}, we show the cost of our evaluation of proprietary models. The cost of input is significantly greater than the cost of output, suggesting that the existing proprietary models still face the problem of excessive token amount when receiving video input.

\setlength{\tabcolsep}{1pt} 
\begin{table}[h]
\centering
\caption{Evaluation Cost of Different Models}
\scalebox{0.76}{
\begin{tabular}{lccc}
\hline
\textbf{Model} & \textbf{Input Cost} & \textbf{Output Cost} & \textbf{Total Cost} \\ \hline
Gemini-1.5-Flash & \$13.41 & \$0.02 & \$13.43 \\ 
Gemini-1.5-Pro & \$223.49 & \$0.29 & \$223.78 \\ 
Gemini-2.0-Flash & \$17.88 & \$0.02 & \$17.90 \\
GPT-4o & \$95.49 & \$0.27 & \$95.77 \\ 
GPT-4o-mini & \$34.31 & \$0.04 & \$34.35 \\ 
Qwen-VL-Max-latest & \$4.31 & \$0.24 & \$4.55 \\ \hline
\end{tabular}
}
\label{tab:evaluation_cost}
\end{table}

\subsection{Prompt}
For each Video-LLM, the input includes an embodied movement video, a single MCQ, and a prompt that introduces the background and output format.

For models such as Gemini-1.5-flash, the videos can be uploaded to cloud storage space, the questions are given to the model separately for evaluation, as is shown in the following prompt:
\begin{tcolorbox}
	\textit{Please assume the role of an aerial agent. The video represents your egocentric observations from the past to the present. Please answer the following questions: \newline <Question> \newline
    <video input>\newline
    The template for the answer is: \newline
Option: [] (Only output one option from 'A' to 'E' here, do not output redundant content) \newline
Reason: [] (Explain why you choose this option) }
\end{tcolorbox}
For models such as GPT-4o and Qwen, the videos are given to the model one by one, together with the all the questions based on this video, and the models are then required to answer all the questions one by one. In this way, we avoid uploading the same video repeatedly, so as to reduce time consumption and token number. Such prompt are shown below:
\begin{tcolorbox}
	\textit{Please assume the role of an aerial agent. The video represents your egocentric observations from the past to the present. Please answer the following questions: \newline <Question 0>, <Question 1>,  ... \newline
	The template for the answer is: \newline
QA0: Option: []; Reason: []. \newline QA1: Option: []; Reason: []... \newline
The Option only outputs one option from 'A' to 'E' here, do not output redundant content. Reason explains why you choose this option.}
\end{tcolorbox}

\section{Fine-Tuning} \label{appendix:Fine-Tuning}
In this experiment, we conducted multimodal fine-tuning on the \texttt{InternVL2-4B} and \texttt{InternVL2-8B} models. During fine-tuning, the visual encoder (\texttt{freeze\_visual\_encoder=True}) and the language model backbone (\texttt{freeze\_llm=True}) were frozen, and only the language model was lightly adapted using LoRA technology (rank 128, alpha 256, dropout 0.05). The experimental data was based on 70\% of the multiple-choice question samples from \texttt{All\_MCQ.jsonl}, with the associated video data totaling 39.56\,GB, stored in the \texttt{video\_LLM/video} directory. The videos were processed by a pre-trained visual encoder to extract spatiotemporal features, which were then concatenated with text embeddings and fed into the model. The maximum input sequence length was set to 8192 to accommodate the joint modeling of long video segments and complex text.

Training employed the AdamW optimizer (learning rate $1\mathrm{e}{-6}$, weight decay 0.05) combined with linear warm-up and cosine annealing scheduling. The batch size per GPU was 4, with a gradient accumulation step of 2, resulting in an effective batch size of 8. The maximum training epoch was set to 1. Video data was efficiently loaded using a distributed data loader, with dynamic mixed precision (\texttt{AmpOptimWrapper}) and gradient clipping (maximum norm 1) enabled. Checkpoints were saved every 64 steps to ensure training stability and storage efficiency.

\section{Error Cases} \label{appendix:Error_Cases}
Table \ref{tab:error_case_1} and Table \ref{tab:error_case_2} shows the results and reasoning process of an Video-LLM (Language and Vision Learning Model) answering questions on a dataset. The table has four columns:

Question Category: This column describes the type of question, such as Action Generation, High-level Planning, etc.

Question: This column contains the specific questions and the corresponding options.

Ground Truth: This column shows the correct answers to the questions, which are carefully checked in human refinement.

Video-LLM Answer: This column provides the answer given by the Video-LLM, and also the reasoning behind the Video-LLM model's choices, demonstrating the model's reasoning process.

Through these questions and their options, answers, and explanations, this table highlights the shortcomings of the Video-LLM in the embodied abilities.

\newpage
\begin{table*}[t]
\centering
\fontsize{6}{7.2}\selectfont
\caption{Error Example Part 1}
\begin{tabular}{|p{2cm}|p{8cm}|p{0.8cm}|p{2.5cm}|}
\hline
\rowcolor[HTML]{548235} \color{white}\textbf{Question Category} & \color{white}\textbf{Question} & \color{white}\textbf{Ground\newline Truth} & \color{white}\textbf{Video-LLM Answer} \\ \hline
Object Recall & According to the video, what is directly above the balcony on the 24th floor? \newline Choices: \newline A. There is a rooftop garden above the balcony on the 24th floor. \newline B. There is a billboard directly above the balcony on the 24th floor. \newline C. There is another balcony directly above the balcony on the 24th floor. \newline D. There is nothing directly above the balcony on the 24th floor. \newline E. There is a drone stationed above the balcony on the 24th floor. & C & Option D; Reason: There is no mention or indication in the video that there is anything directly above the balcony on the 24th floor. \\ \hline
Object Recall & Question: In the video, what is on the left side of the ""Huaji Stone Rolled Rice Rice Roll"" shop?\newline Choices: A. Large parking lot full of cars.\newline B. Another store with a green storefront and a plus sign that says ""Dashenlin"".\newline C. A shop with a black car parked at the door, and the signboard reads ""NOBEE Nobeni"".\newline D. A small, unpaved area with some overgrown vegetation.\newline E. A bus stop with shelters and benches.
 & C & Option B; Reason: The video shows that on the left side of the "Hugj Stone Rolled Rice Roll" shop, there is another store with a green storefront and a plus sign that says "Dashenlin". \\ \hline
Duration & Which takes longer, the UAV's descent towards the street level or its maneuvering between the buildings? \newline Choices: \newline A. The UAV's descent towards the street level takes longer than its maneuvering between the buildings. \newline B. The UAV's descent towards the street level takes shorter than its maneuvering between the buildings. \newline C. The UAV's descent towards the street level takes the same time as its maneuvering between the buildings. & A & Option B; Reason: The UAV's descent towards the street level is a straightforward movement, while maneuvering between buildings requires more complex navigation and adjustments, which would take longer. \\ \hline
Duration & Which takes longer, the UAV's descent to street level near the bus stop or its flight over the center of the street between skyscrapers? \newline Choices: \newline A. The UAV's descent to the street level takes longer than its flight over the center of the street. \newline B. The UAV's descent to the street level takes shorter than its flight over the center of the street. \newline C. The UAV's descent to the street level takes the same time as its flight over the center of the street. & A & Option B; Reason: The video shows that UAV's flight over the center of the street between skyscrapers takes longer than its descent to the street level near the bus stop. \\ \hline
Trajectory Captioning & According to the video, which of the following sentences better summarizes the movement of the drone? \newline Choices: \newline A. The drone ascends from the facade of the high-rise building, moves forward, turns right on the last side and descends, and reaches the top of the seaside fence. \newline B. The drone descends from the city center to street level and crosses the main thoroughfare to the top of a skyscraper. \newline C. The drone flew from an open street next to a building over several rooftops to a park near the ocean. \newline D. The drone travels from a bridge over the river through numerous skyscrapers to the main city park. \newline E. The drone turns left from above the road and follows the path along the water into the forest area. & A & Option D; Reason: The video shows the drone starting from the facade of a tall building, moving forward, turning right on the last side, and descending to reach the top of the seaside fence. This matches option D. \\ \hline
Start/End Position & Question: According to the video, where are the starting and ending points of the drone’s journey? \newline Choice: A. The drone starts at a large intersection on a city street and ends at a store window with a blue canopy. \newline B. The drone starts above the street and ends at a shop window with a white canopy. \newline C. The drone starts from the fire exit in front of the building and ends at the subway entrance. \newline D. The drone starts from a fire escape on the middle floor of the building and ends at a store window with a white canopy. \newline E. The drone starts in the center of the street and ends in the park. & B & Option D; Reason: The video shows the drone starting from a fire escape on the middle floor of the building and ending at a store window with a white canopy. \\ \hline
Proximity & Question: After the drone moves to the outside of the park, how does the distance between the drone and the staircase-like building change? \newline Choices: \newline A. The distance between the drone and the staircase-like building decreases. \newline B. The distance between the drone and the staircase-like building increases. \newline C. The distance between the drone and the staircase-like building remains constant. & A & Option B; Reason: As the drone moves to the outside of the park, it appears to be moving away from the staircase-like building, increasing the distance between them. \\ \hline
Scene Recall & Question: What objects did the drone observe when it hovered over the trees in the middle of the intersection? \newline Choices: \newline A. The drone observes road markings and phone booths as it hovers over the woods. \newline B. The drone observes a bridge, a car, and a lamppost as it hovers over the woods. \newline C. The drone observed skyscrapers, rooftop gardens, and flags while hovering over trees. \newline D. The drone observed a playground, fountain, and street vendors as it hovered over the woods. \newline E. The drone observed a church, a bell tower and a water tower while hovering over the woods. & A & Option C; Reason: When the drone hovers over the trees in the middle of the intersection, it observes skyscrapers, rooftop gardens, and flags. \\ \hline
Counterfactual & Question: If the drone does not rise and move parallel to the group of skyscrapers, but chooses to fly directly over the roofs of lower buildings, can it still reach the trees in the middle of the zebra crossing at the intersection? \newline Choices: \newline A. If the drone chooses to fly directly over the rooftop, it can complete the mission since it has a direct path, but it will take longer. \newline B. If the drone chooses to fly directly over the roof, it will not be able to complete its mission because it will miss the intersection. \newline C. If the drone chooses to fly directly over the roof, it can complete the mission because the alternative path is more efficient. & A & Option C; Reason: If the drone chooses to fly directly over the roofs of lower buildings, it can still reach the trees in the middle of the zebra crossing at the intersection because this alternative path is more efficient and direct. \\ \hline
Counterfactual & Question: If the drone does not fly along the winding road over the lake, but flies directly over the open fields, can it still reach the narrow bridge across the lake in the park? \newline Choices: \newline A. If the drone chooses to fly directly over an open field, it can accomplish the mission because the field provides a clear path, whereas the alternative would require less time. \newline B. If the drone chooses to fly directly over the open field, it will not be able to complete its mission because it will get lost in the middle. \newline C. If the drone chooses to fly directly over the open field, it can complete the mission because the alternative path is more efficient. & B & Option C; Reason: Flying directly over an open field can help the drone avoid trees, making the task easier and potentially faster. \\ \hline
Sequence Recall & Question: What is the next step after the drone crosses the second intersection? \newline Choice: \newline A. The drone ascends to gain a higher vantage point. \newline B. The drone continues its progress over the roof. \newline C. The drone descends close to the roof. \newline D. The drone turns sharply to the right. \newline E. The drone hovers in place. & C & Option: B; Reason: After crossing the second intersection, the drone continues its journey over the roof, as seen in the video. \\ \hline
\end{tabular}
\label{tab:error_case_1}
\end{table*}

\begin{table*}[t]
\centering
\fontsize{6}{7.2}\selectfont
\caption{Error Example Part 2}
\begin{tabular}{|p{2cm}|p{8cm}|p{0.8cm}|p{2.5cm}|}
\hline
\rowcolor[HTML]{548235} \color{white}\textbf{Question Category} & \color{white}\textbf{Question} & \color{white}\textbf{Ground\newline Truth} & \color{white}\textbf{Video-LLM Answer} \\ \hline
Causal & Question: Why do drones keep rising after leaving the beach? \newline Choice: \newline A. The drone ascends to capture a wider view of the waterfront. \newline B. The drone rises, avoiding trees and lamp posts. \newline C. The drone rises to avoid hitting the roof of the building. \newline D. The drone ascends to the upper level of the building. \newline E. The drone ascends to the rooftop helipad. & A & Option: C; Reason: The drone rises to align with the upper level of the building, as shown in the video frames. \\ \hline
Landmark Position & Question: The drone is navigating to (the parking space on the right after entering the community entrance). When the drone reaches the current location, what is the direction to the destination? \newline Choice: \newline A. The drone is currently located above the parking space on the right and has just entered the entrance of the community. \newline B. The drone is currently located far away from the community, above a wide avenue. \newline C. The drone is currently located above the community center and not close enough to the parking space. \newline D. The drone is currently located above the left side of the community, near another high-rise building. \newline E. The drone is currently located above the community center and not close enough to the parking space. & A & Option: E; Reason: The drone is currently located over the community center and needs to move towards the parking space on the right. \\ \hline
Goal Detection & Question: The drone is navigating to the balcony on the 13th floor. Is the destination within sight at the current location? If yes, where is the destination in the view? \newline Choice: \newline A. The balcony on the 13th floor is currently located below the center of the drone's field of view. \newline B. The balcony on the 13th floor is currently located in the upper left corner of the drone's field of view. \newline C. The balcony on the 13th floor is currently located on the right side of the drone's field of view. \newline D. The balcony on the 13th floor is currently located in the lower right corner of the drone's field of view. & A & Option: D; Reason: The video does not show the balcony on the 13th floor within the drone's field of view at the current location. \\ \hline
Goal Detection & Question: The drone is navigating to the balcony on the 13th floor. Is the destination within sight at the current location? If yes, where is the destination in the view? \newline Choice: \newline A. The balcony on the 13th floor is currently located below the center of the drone's field of view. \newline B. The balcony on the 13th floor is currently located in the upper left corner of the drone's field of view. \newline C. The balcony on the 13th floor is currently located on the right side of the drone's field of view. \newline D. The balcony on the 13th floor is currently located in the lower right corner of the drone's field of view. & A & Option: D; Reason: The video does not show the balcony on the 13th floor within the drone's field of view at the current location. \\ \hline
Action Generation & The drone is navigating to the 31st floor balcony. What is the next move for the drone, given the target location for the drone's next move? \newline \textbf{Choices:} \newline A. Rise. \newline B. Descend. \newline C. Fly forward. \newline D. Turn left. \newline E. Turn right. \newline F. Rotate the camera upward. \newline G. Rotate the camera downward. & B & Option: A; Reason: The drone needs to rise to reach the 31st floor balcony, so the next move should be to ascend. This aligns with option A. \\ \hline
Action Generation & The drone is navigating to the 13th floor balcony. What is the next move for the drone, given the target location for the drone's next move? \newline \textbf{Choices:} \newline A. Rise. \newline B. Descend. \newline C. Fly forward. \newline D. Turn left after descending. \newline E. Fly forward. \newline F. Turn left. \newline G. Turn right. \newline H. Rotate the camera upward. \newline I. Rotate the camera downward. & B & Option: F; Reason: The drone needs to rotate the camera upward to align with the balcony on the 13th floor. \\ \hline

High-level Planning & The drone is navigating to a nearby bus stop. Thinking back on past actions and objects seen in the video, what should the drone approach next in order to get from its current location to its destination? \newline \textbf{Choices:} \newline A. The center of the square. \newline B. The top of the tallest building in the distance. \newline C. Crossing the street. \newline D. The end of the parking lot. \newline E. Street edge. & E & Option: C; Reason: Crossing the street is the next step for the drone to get to the bus stop. \\ \hline
High-level Planning & The drone is navigating to the largest intersection nearby. Thinking back on past actions and objects seen in the video, what should the drone approach next in order to get from its current location to its destination? \newline \textbf{Choices:} \newline A. The top of the traffic light at the intersection. \newline B. Center of the roundabout. \newline C. The highest point of the residential building across the street. \newline D. Street parking. \newline E. The tallest building visible on the horizon. & B & Option: A; Reason: The top of the traffic light at the intersection. To navigate to the intersection, approaching the top of the traffic light would provide a clear vantage point for identifying the intersection. \\ \hline
Progress Evaluation & The drone moves according to a series of movement instructions. What are the drones doing now? \newline \textbf{Choices:} \newline A. Fly towards the sun. \newline B. Go to the clear between the buildings. \newline C. Go to the river. \newline D. Go to the brown building. \newline E. Go to the sun. & B & Option: C; Reason: The drone is currently positioned to the right of the trees as it moves through the cityscape. \\ \hline
Progress Evaluation & The drone moves according to a series of movement instructions. What are the drones doing now? \newline \textbf{Choices:} \newline A. Fly towards the sun. \newline B. Turn right to the residential building. \newline C. Increase the height of the drone. \newline D. Change the camera gimbal from downward to normal. \newline E. Fly towards the road. & B & Option: D; Reason: The drone is currently capturing a downward view of the city, and changing the camera gimbal to normal will provide a more standard perspective.  \\ \hline

\end{tabular}
\label{tab:error_case_2}
\end{table*}